\documentclass[10pt,twocolumn,letterpaper]{article}

\usepackage[pagenumbers]{cvpr} 

\usepackage[dvipsnames]{xcolor}

\definecolor{cvprblue}{rgb}{0.21,0.49,0.74}
\usepackage[pagebackref,breaklinks,colorlinks,citecolor=cvprblue,urlcolor = black]{hyperref}
\usepackage{diagbox}
\usepackage{amsthm}
\usepackage{calligra}
\usepackage{multirow}
\usepackage{algorithm}
\usepackage{algorithmic}
\usepackage[numbers]{natbib}
\usepackage{makecell}
\usepackage{subcaption}
\usepackage[dvipsnames]{xcolor}
\usepackage{titletoc}
\theoremstyle{definition}
\newtheorem{precondition}{Precondition}

\theoremstyle{definition}
\newtheorem{definition}{Definition}

\definecolor{gray}{rgb}{0.5,0.5,0.5}
\definecolor{RedColor}{RGB}{255, 0, 84} 
\definecolor{BlueColor}{RGB}{5, 191, 219} 
\definecolor{GreenColor}{RGB}{22, 204, 0} 
\definecolor{YellowColor}{RGB}{225, 168, 0} 
\definecolor{PurpleColor}{RGB}{204, 51, 255} 

\newcommand{\added}[1]{{#1}}

\newlength\myindent
\newcommand\bindent{%
  \begingroup
  \setlength{\itemindent}{\myindent}
  \addtolength{\algorithmicindent}{\myindent}
}
\newcommand\eindent{\endgroup}
\newcommand*\samethanks[1][\value{footnote}]{\footnotemark[#1]}

\titlecontents{section}
[3em] 
{\addvspace{5pt}} 
{\contentslabel{2em}} 
{} 
{\titlerule*[1pc]{.}\contentspage} 

\titlecontents{subsection}
[5em] 
{\addvspace{5pt}}
{\contentslabel{2.6em}} 
{}
{\titlerule*[1pc]{.}\contentspage} 

\def\paperID{***} 
\def\confName{CVPR}
\def\confYear{2024}

\title{Dancing with Still Images: \\Video Distillation via Static-Dynamic Disentanglement}

\author{
  Ziyu Wang\thanks{The first two authors contribute equally.}~~~~~Yue Xu\samethanks~~~~~Cewu Lu~~~~~Yong-Lu Li\thanks{Corresponding author.}\\
Shanghai Jiao Tong University\\
{\tt\small \{wangxiaoyi2021, silicxuyue, lucewu, yonglu\_li\}@sjtu.edu.cn}
}

\begin{document}

\maketitle

\begin{abstract}

Recently, dataset distillation has paved the way towards efficient machine learning, especially for image datasets. However, the distillation for videos, characterized by an exclusive temporal dimension, remains an underexplored domain. 
In this work, we provide the first systematic study of video distillation and introduce a taxonomy to categorize temporal compression. 
Our investigation reveals that the temporal information is usually not well learned during distillation, and the temporal dimension of synthetic data contributes little. 
The observations motivate our unified framework of disentangling the dynamic and static information in the videos. It first distills the videos into still images as static memory and then compensates the dynamic and motion information with a learnable dynamic memory block.
Our method achieves state-of-the-art on video datasets at different scales, with a notably smaller memory storage budget.
\textbf{\added{Our code is available at} \href{https://github.com/yuz1wan/video_distillation}{https://github.com/yuz1wan/video\_distillation}}.
\end{abstract}
    
\section{Introduction}
\label{sec:intro}
Dataset distillation, as an emerging direction recently, compresses the original dataset into a smaller one while maintaining training effectiveness. 
It alleviates the challenges of costly training due to the increasingly large datasets and models.
It is widely adopted in various downstream fields including federated learning and continual learning.

Recent works on dataset distillation mainly focus on distilling images~\cite{DD,DC,DM,MTT,FRePo,KIP,KIP2,RFAD,DSA,IDC,tesla,CAFE,haba,linba}. 
Though some methods seem to seamlessly adapt to other data formats or modalities~\cite{li2021data-TEXT, wu2023multimodal-CLIP, xu2023kernel-GRAPH}, few works studied video distillation.
Compared to image data, videos possess an additional temporal dimension, which significantly adds to the time and space complexity of the distillation algorithms and is already hardly affordable when the instance-per-class (IPC) is large. Besides, the scale of video datasets~\cite{ucf101,hmdb51,kinetics} is usually more intimidating.
However, the high temporal redundancy of videos is very suitable for and can be well exploited by dataset distillation methods, providing a good opportunity for dataset distillation. 
Therefore, in this work, we firstly and systematically study the dataset distillation on video data, especially involving the compression of the video temporal redundancy. 

\begin{figure}[t]
    \centering
    \includegraphics[width=0.99\linewidth]{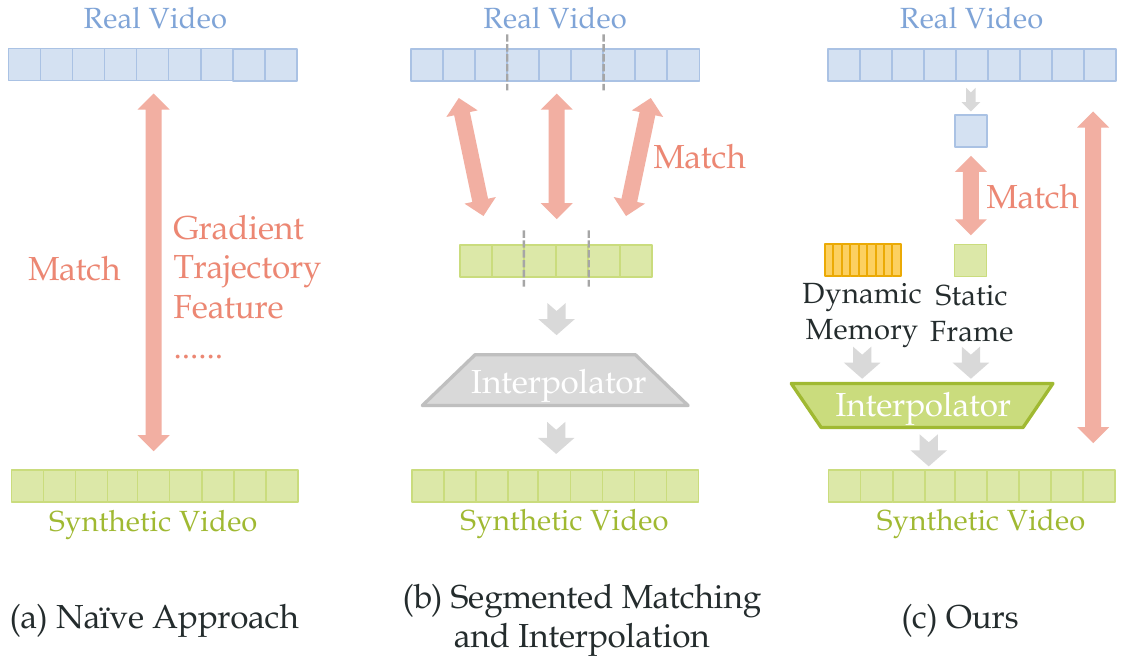}
    \vspace{-5px}
    \caption{(a) Naive video distillation methods simply match the training dynamics (gradient, feature, trajectory, \etc) of the real and synthetic videos. (b) To exploit the temporal redundancy of videos, we propose a paradigm with segmented matching and interpolation techniques to cover all levels of temporal condensation. 
    (c) Based on this paradigm and our observations, we propose an approach of efficient static frame distillation and motion compensation, with better efficiency and performance.}
    \vspace{-7px}
    \label{fig:teaser}
\end{figure}

Currently, dataset distillation approaches directly align the training dynamics (gradient~\cite{DC}, trajectory~\cite{MTT}, feature~\cite{DM}, \etc.) of real and synthetic. On video datasets, these methods simply match all the real and synthetic frames and the frame correspondence can be depicted as a complete bipartite graph.
Therefore, to condense the temporal dimension, we can either reduce the synthetic frames or prune the bipartite correspondence graph between real and synthetic frames.
For the further analysis of temporal condensation in video distillation, we put these two schemes under one unified framework, namely ``segmented matching and interpolation'' (Fig.~\ref{fig:teaser}(b)): the real videos are synthetic videos are partitioned to multiple segments and distillation are applied within each real-synthetic segments pair; the synthetic videos are then interpolated to the required video length. This framework can cover most scenarios at different levels of temporal condensation.

Then, to thoroughly study temporal condensation, we build a taxonomy for various temporal correspondences and classify the potential methods based on four essential dimensions: the numbers of synthetic frames and real frames for matching, the number of segments, and the interpolation algorithm.
Along these four dimensions, we conduct comparisons of the distillation performance and computation cost. 
Empirical analysis shows that, though increasing frame numbers does enhance distillation performance, the improvement is marginal and comes at the expense of considerably longer training times and higher costs; and frame segmentation reduces the training costs, but brings a substantial decrease in model performance.
These important taxonomies and observations could guide the research and the efficient algorithm design of video distillation.

In \added{light} of the above observation, we propose a unified framework for video distillation to exploit the unique temporal redundancy of videos.
Our observation implies that dense frame correspondence is non-critical in the video distillation task. Hence, to maximize the efficiency, we reduce the real frames, synthetic frames, and segment length to 1. This is equivalent to image distillation with which we can distill the \textit{static information} in the first stage.
Second, we use a learnable \textit{dynamic memory} to compensate for the loss of dynamic information. The static and dynamic memories are then combined with an integrator network.
\added{Our paradigm can be effortlessly applied to existing algorithms to enhance performance with a memory storage budget (referred to as \textit{storage} here) used.
We embed our method to various data distillation algorithms including DM~\cite{DM}, MTT~\cite{MTT}, FRePo~\cite{FRePo}, and achieve state-of-the-art with less storage. 
Our approach could achieve comparable performance with $<$50$\%$ memory storage budget and bring substantial improvement with a comparable one. }

Overall, our contributions are:
(1) We propose the very first work that systematically studies the video dataset distillation.
(2) We introduce a novel taxonomy for temporal condensation in video distillation methods, which guides our and future works.
(3) We propose a novel paradigm, enabling existing image distillation techniques to achieve improved results when applied to video distillation while using an even smaller memory storage budget. 
\section{Related work}
\label{sec:related}
{\bf Dataset Distillation/Condensation.}
Dataset distillation~\cite{DD}, endeavors to condense large datasets into smaller ones while maintaining comparable training performance. The algorithms fall into the following categories:

(1) \underline{\textit{Performance Matching}}: following the very first work of DD~\cite{DD}, a broad category of techniques employs bi-level optimization. 
A few methods integrate kernel ridge regression (KRR) to reduce the computational complexity of bi-level optimization, where KIP~\cite{KIP, KIP2} employs the Neural Tangents Kernel (NTK), while RFAD~\cite{RFAD} adopts the Empirical Neural Network Gaussian Process (NNGP). FRePo~\cite{FRePo} separates a neural network into a feature extractor and a linear classifier to optimize.

(2) \underline{\textit{Parameter Matching}}:
DC~\cite{DC} aligns the single-step gradients of synthetic and real data. In line with DC, DSA~\cite{DSA} enhances this approach through symmetrical image augmentation, and IDC~\cite{IDC} enhances by storing synthetic data in lower resolutions. MTT~\cite{MTT} first applies multi-step parameter matching, and TESLA~\cite{tesla} reduces memory usage and uses learnable soft labels.

(3) \underline{\textit{Distribution Matching}}:
DM~\cite{DM} directly aligns the features of real and synthetic data, while CAFE~\cite{CAFE} ensures statistical feature properties from synthetic and real samples are consistent across all network layers except the final one.

(4) \underline{\textit{Factorization Methods}} decompress the full dataset into two components: base and hallucinator. HaBa~\cite{haba} employs task-input-like bases and ConvNet hallucinators, while LinBa~\cite{linba} integrates a linear hallucinator with given predictands.
Inspired by these methods, we factorize the static and dynamic information in video distillation to minimize temporal redundancy and reduce storage costs.

\noindent{\bf Video Recognition.}
Video recognition involves the classification of videos into semantic classes, \eg, human actions and scenes.
Currently, 
there are several main design philosophies for video recognition with deep learning: 
(1) \underline{\textit{2D Convolution-Based}}:
The most intuitive approach is to break down the video into individual frames and process the frames individually. Then temporal aggregation (pooling, LSTM, GRU, \etc) is used for getting video-level features, which is then used for classification.
(2) \underline{\textit{3D Convolution-Based}}.
To adopt early aggregation of temporal features, 
the presence of an additional temporal dimension in videos naturally suggests the possibility of employing 3D convolutional networks. 
\citet{c3d} propose C3D, and then \citet{kinetics} extend the pre-trained models of 2D convolutional networks to 3D, with consideration of alleviating the challenges posed by the large number of parameters in 3D convolutions. 
Meanwhile, efforts are dedicated to low-rank approximations for 3D convolutions. R(2+1)D~\cite{r2+1d} employs a spatial 2D convolutional structure combined with a temporal 1D convolution to achieve pseudo-3D convolutions.
(3) \underline{\textit{Transformer-Based}}.
With the success of attention mechanisms in natural language processing, 
the long-range effectiveness of self-attention determines its suitability for video recognition. Therefore, there have also emerged some models~\cite{actionclip,timesformer,videovae} for videos based on Transformers.

\section{Pre-analysis}
\label{sec:taxonomy}

The temporal redundancy has been widely discussed for video understanding, while we focus on analyzing the temporal compression in dataset distillation. In this section, we first propose some basic principles for temporal compression (Sec.~\ref{subsec:preliminary}). We use a generic paradigm to describe the temporal compression strategies and further propose a taxonomy of compression according to four factors (Sec.~\ref{subsec:seg-comp-interp}), along with which we conduct comprehensively comparisons and obtain our observations and conclusions, supporting further study on video distillation (Sec.~\ref{subsec:compare-compression}).

\subsection{Preliminaries}
\label{subsec:preliminary}

{\bf Video Temporal Redundancy.}
Most current works are dedicated to developing and improving methods for compressing image datasets in terms of quantity. 
Video data have an additional temporal dimension as its major difference from images, suggesting the presence of temporal redundancy. 
Video temporal redundancy study has a long history~\cite{nflb,wmvv}. 
Researchers have long observed the significant temporal redundancy, which has diverse causes. For example, videos inherently exhibit substantial similarity between adjacent frames, leading to low temporal information utilization during data usage.  
Thus, we focus on, analyze, and exploit the temporal compression for video distillation.

{\bf Precondition for Temporal Correspondence.}
To study the temporal compression, we are categorizing the temporal correspondence for the distillation matching (\eg gradient/trajectory/distribution matching) between real and synthetic videos.
We can put all scenarios into one formulation, which covers all possible methods from the temporal aspect:

\begin{definition}
    \textbf{Compressed video distillation} involves real video $R$ and synthetic video $S$ with asymmetrical lengths. Multiple frame sequences are drawn from $R$, $S$ and paired: $(R_1, S_1)$, $(R_2, S_2),~\cdots,~(R_K, S_K)$, where $R_i\subseteq R$, $S_i\subseteq S$, $\forall i=1,\cdots,K$. Distillation is to apply matching algorithms to each frame sequence pair individually.
\end{definition}

\begin{figure}[t]
    \centering
    \includegraphics[width=0.99\linewidth]{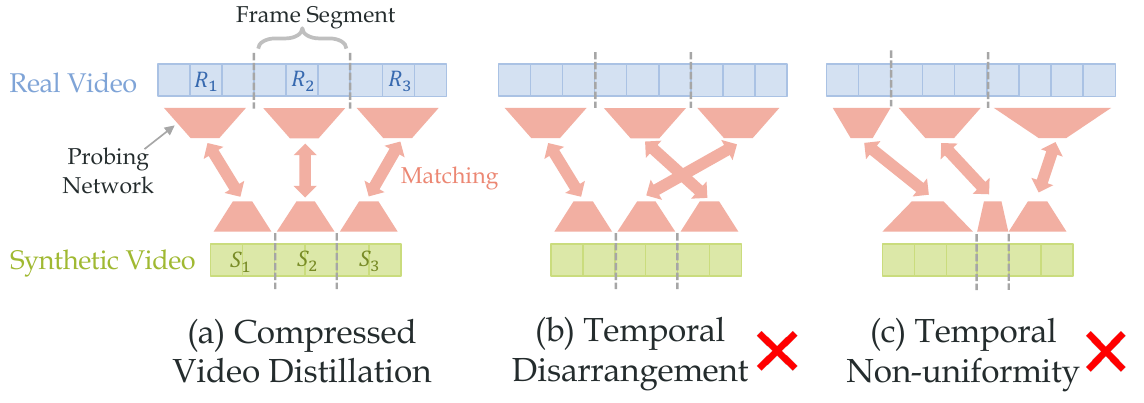}
    \vspace{-6px}
    \caption{The distillation setting in (a) obeys temporal consistency, while (b) and (c) violate the two consistency preconditions.}
    \vspace{-8px}
    \label{fig:temporal-consistency}
\end{figure}

Fig.~\ref{fig:temporal-consistency}(a) gives an example of compressed video distillation.
Though the existing distillation methods usually produce irregular and ``freeform'' patterns that are cryptic for humans, on video data, we intuitively desire an algorithm that obeys \textbf{temporal consistency}. 
Otherwise, the video algorithm could degrade due to the loss of correct temporal information.
The temporal consistency is ensured by two preconditions: the real and synthetic frames sampled for distillation should be in the correct order and follow a uniform flow of time. 
More specifically, the orderedness is:

\begin{precondition}
    \textbf{(Orderedness)} 
    During the compressed video distillation of real video $R$ and synthetic video $S$, given two frame sequence pairs $(R_1, S_1)$, $(R_2, S_2)$,
    we define a partial order $\leq_T$: $(R_1, S_1)\leq(R_2, S_2)$ \textit{iff} any frame in $R_1$ occurs earlier than any frame in $R_2$ and any frame in $S_1$ occurs earlier than any frame in $S_2$.
    The distillation process obeys \textbf{orderedness} \textit{iff} all the frame sequence pairs for $R$ and $S$ yield a total ordering associated with $\leq_T$.
\end{precondition}

That is, all frame sequence pairs are comparable and associated with $\leq_T$ and both real and synthetic sequences are in the correct time order. The matching strategy in Fig.~\ref{fig:temporal-consistency}(b) does not meet orderedness as the two matching pairs at right are in the wrong temporal order.

\begin{precondition}
    \textbf{(Uniformity)} Compressed video distillation of real video $R$ and synthetic video $S$, obeys \textbf{uniformity} \textit{iff} $|R_1|=|R_2|=\cdots=|R_K|$, $|S_1|=\cdots=|S_K|$.
\end{precondition}

All the real frame sequences have the same length, as the same to synthetic frame sequences. This ensures that the synthetic video we learn follows a uniform flow of time, \eg, the matching strategy in Fig.~\ref{fig:temporal-consistency}(c) leads to non-uniform FPS, and the synthetic frames at left will learn a much smaller FPS than the frames at right. \added{We also justify the uniformity with experiments in the supplementary.}

\begin{figure}[t]
    \centering
    \includegraphics[width=0.99\linewidth]{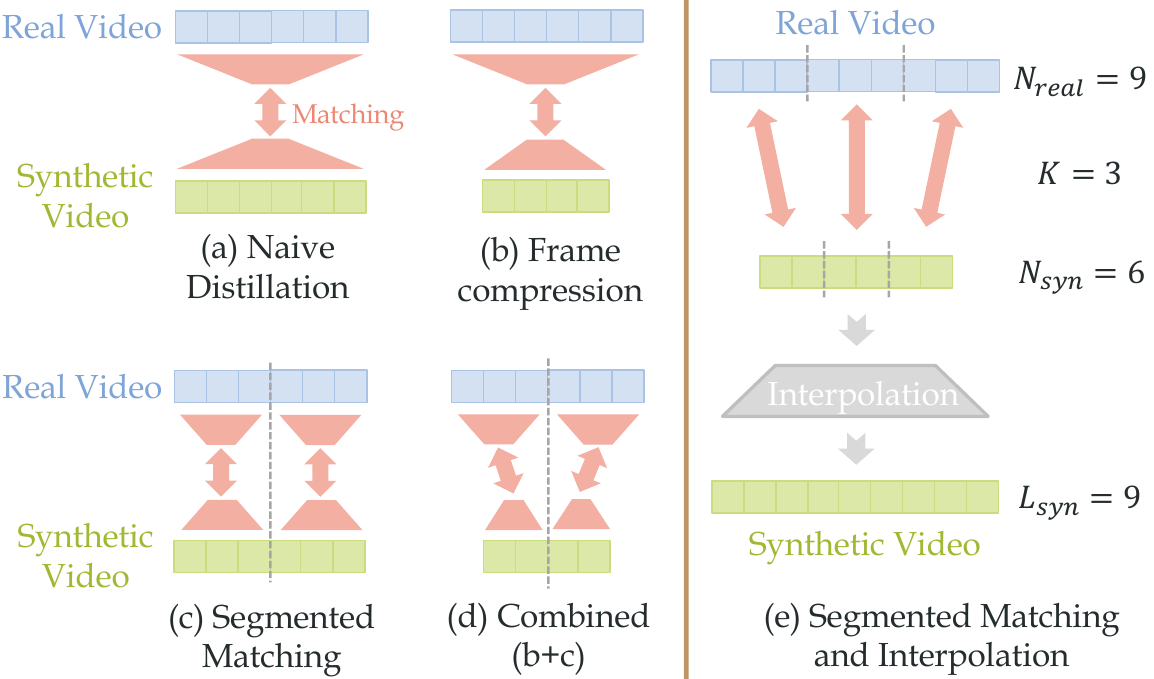}
    \vspace{-5px}
    \caption{\textbf{Left}: Different types of video distillation that obey temporal consistency. \textbf{Right}: A basic framework for compressed video distillation, which condenses the temporal dimension by distillation, and interpolates the synthetic frames to the target length. }
    \vspace{-7px}
    \label{fig:frame-corresp}
\end{figure}

These two preconditions narrow the searching space of temporal condensation strategies, enabling us to systematically categorize and analyze the compressed video distillation.
In Fig.~\ref{fig:frame-corresp}(a), the naive algorithm may match all synthetic frames to real frames. To compress the temporal dimension, we can either reduce the number of frames (Fig.~\ref{fig:frame-corresp}(b)) or reduce the correspondence by temporal segmentation (Fig.~\ref{fig:frame-corresp}(c)).
Note that we do not force orderedness and uniformity \textit{within} each frame sequence pair for matching, since the distillation matching algorithms themselves could drive the synthetic data to follow the consistency.

\subsection{Segmented Matching and Interpolation}
\label{subsec:seg-comp-interp}

To take a step further, we propose \textbf{Segmented Matching and Interpolation} framework for the quantitative comparison and analysis of temporal condensation (Fig.~\ref{fig:frame-corresp}(e)).
Given a target synthetic frame number $L_{syn}$, to approach flexible compression rate, we distill $N_{real}$ real frames to $N_{syn}$ frames and interpolate the frames to our target $L_{syn}$.
Specifically, the real video and synthetic video are \textit{segmented} evenly for the pairwise distillation matching, which is the only valid strategy for temporal consistency. The interpolation enables the compression of synthetic video.
This paradigm covers most video distillation methods from the perspective of temporal compression, and the extent of compression can be parameterized by the four factors:

{\bf (1) Number of Independent Synthetic Frames ($N_{syn}$)} is the size of the trainable synthetic frames. These frames are learned with dataset distillation algorithms and will be interpolated to the target length of synthetic video $L_s$. Smaller $N_{syn}$ indicates a larger temporal compression rate.

{\bf (2) Number of Real Frames ($N_{real}$)} is the size of real frames for distillation matching. Larger $N_{real}$ implies a larger receptive field for synthetic video during distillation.

{\bf (3) Number of Segments ($K$)}. We cut the real and synthetic videos into the same number of segments and apply distillation between the pairs of real and synthetic segments. Larger $K$ could reduce the training time since the distillation algorithm receives smaller segments with fewer frames.

{\bf (4) Interpolation Algorithm ($\mathcal{I}$)} interpolates the $N_{syn}$ independent synthetic frames to the required synthetic video length. Our algorithms are detailed in Sec.~\ref{subsec:compare-compression}.

\begin{figure}[t]
    \centering
    \includegraphics[width=0.99\linewidth]{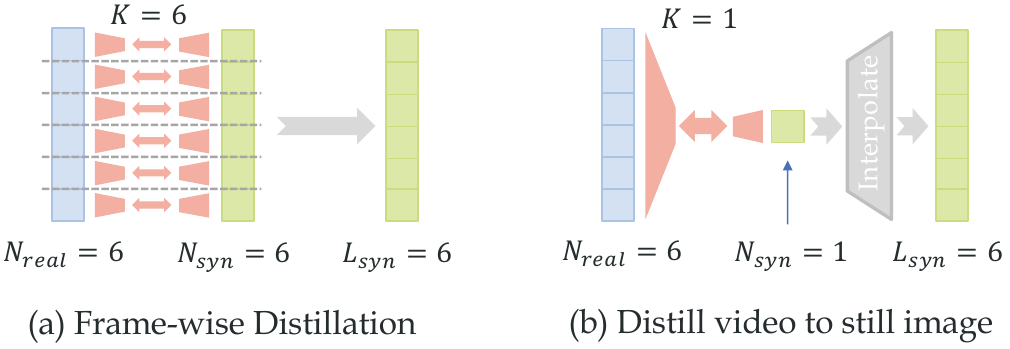}
    \vspace{-5px}
    \caption{Examples with different compression levels. (a) use an image distillation algorithm to distill the frames one by one. (b) distills the video into a single image.}
    \label{fig:compress-examples}
    \vspace{-7px}
\end{figure}

So the level of temporal compression can be uniquely determined by a quadruplet $(N_{syn}, N_{real}, K, \mathcal{I})$. We show some examples in Fig.~\ref{fig:compress-examples} with different combinations. 
In the following section, we will compare and analyze these four axes and put forward some empirical conclusions to drive future studies on video distillation.

\begin{figure*}[t]
  \centering
  \begin{subfigure}{0.345\textwidth}
    \includegraphics[width=\linewidth]{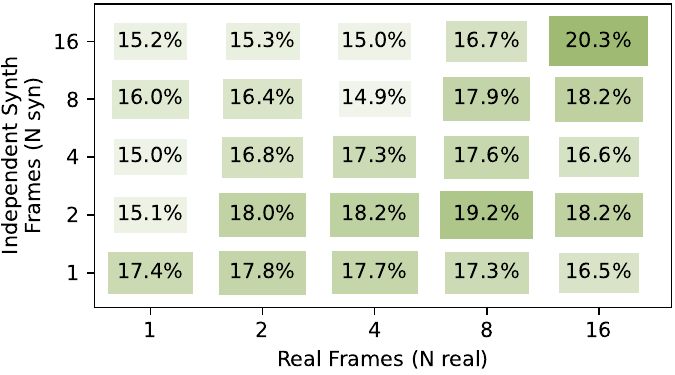}
    \caption{Distillation accuracy.}
  \end{subfigure}
  \begin{subfigure}{0.305\textwidth}
    \includegraphics[width=\linewidth]{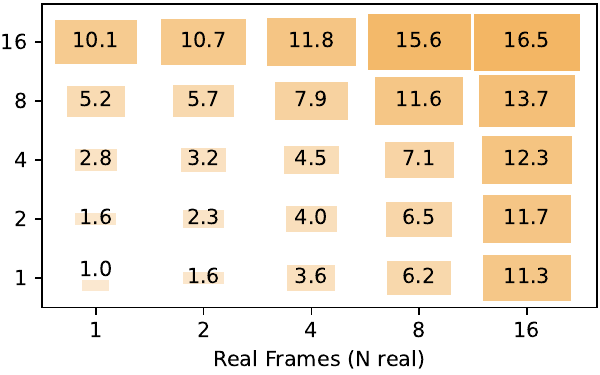}
    \caption{Memory consumption (GB).}
  \end{subfigure}
  \begin{subfigure}{0.305\textwidth}
    \includegraphics[width=\linewidth]{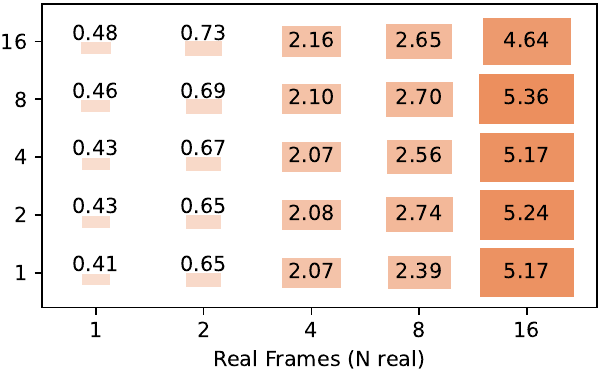}
    \caption{Training time (seconds per iteration).}
  \end{subfigure}
  \caption{Model performance (a) and efficiency (b, c) comparison with different independent synthetic frames and real frames number $N_{syn}$,  $N_{real}$, with DM~\cite{DM} and ConvNet+GRU.}
  \vspace{-5px}
  \label{fig:nreal-nsyn-compare}
\end{figure*}

\begin{figure*}[t]
  \centering
  \begin{subfigure}{0.340\textwidth}
    \includegraphics[width=\linewidth]{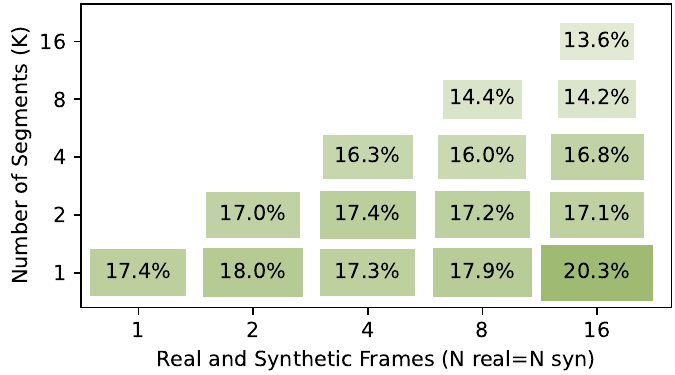}
    \caption{Distillation accuracy.}
  \end{subfigure}
  \begin{subfigure}{0.305\textwidth}
    \includegraphics[width=\linewidth]{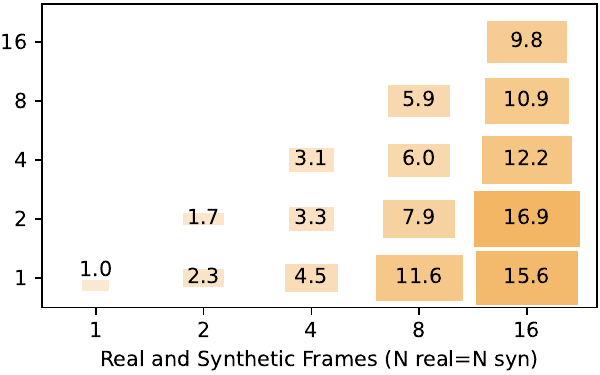}
    \caption{Memory consumption (GB).}
  \end{subfigure}
  \begin{subfigure}{0.305\textwidth}
    \includegraphics[width=\linewidth]{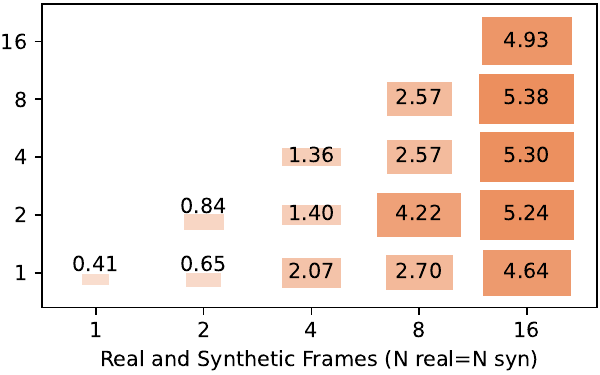}
    \caption{Training time (seconds per iteration).}
  \end{subfigure}
  \caption{Performance (a) and efficiency (b, c) comparison with different numbers of segments $K$, with DM~\cite{DM} and ConvNet+GRU.}
  \vspace{-7px}
  \label{fig:segment-compare}
\end{figure*}

\subsection{Comparison of Temporal Compression}
\label{subsec:compare-compression}

To investigate the effects of different temporal compression levels, we conduct comprehensive experiments with different $N_{syn}$, $N_{real}$, $K$, and $\mathcal{I}$. In the following experiments, the DM~\cite{DM} algorithm is adopted and we use ConvNet (3-layer convolutional network from \cite{DC}) with one layer GRU head~\cite{GRU} as our backbone network.

{\bf Comparison of $N_{syn}$ and $N_{real}$ values.}
We first compare the models with different $N_{syn}$ (number of independent synthetic frames) and $N_{real}$ (number of real frames) and visualize in Fig.~\ref{fig:nreal-nsyn-compare}. We implement the segmented matching segment number $K=1$.
The performance is Fig.~\ref{fig:nreal-nsyn-compare}(a) shows that: (1) distilling the video to still image could yield decent accuracy (over 17\%), (2) larger $N_{real}$ and $N_{syn}$ brings \textit{minor} performance gain (less than 3\%), (3) the model degrades when $N_{real}<N_{syn}$ (left-top of the figure), potentially due to insufficient temporal information in the real data.
Fig.~\ref{fig:nreal-nsyn-compare}(b) and (c) show that larger $N_{real}$ and $N_{syn}$ lead to significant memory consumption, \eg model with $N_{real}=N_{syn}=16$ takes $16\times$ GPU memory than a $N_{real}=N_{syn}=1$ model. And models with large $N_{real}$ also take more training time.
\added{And we can also read Fig.~\ref{fig:nreal-nsyn-compare} diagonally to fix the per-frame receptive field $N_{real}/N_{syn}$, and basically larger ratio leads to better performance.}

{\bf Comparison of $K$ values (segments).}
We study the effects of $K$ in Fig.~\ref{fig:segment-compare}. The segmentation could notably reduce memory consumption (up to 40\%) while maintaining the training speed. However, the efficiency is achieved at the cost of performance as the segmentation decreases the ``receptive field'' of each synthetic frame.

{\bf Comparison of Interpolation Algorithms $\mathcal{I}$.}
$\mathcal{I}$ is critical to compressed video distillation, especially when $N_{syn}$ is small. We use various interpolation methods:
{\bf (1) Duplication}
is simply copying the learned synthetic frames to the required length, or namely \textit{nearest interpolation}.
\eg a 2-frame video $\left[f_1, f_2\right]$ can be interpolated to 4-frame video $\left[f_1, f_1, f_2, f_2\right]$.
{\bf (2) Linear interpolation} generates intermediate frames by blending adjacent reference frames. The frame $f_t$ at time $t$ will be the weighted sum of nearest reference frames $f_{t_1}$, $f_{t_2}$ according to their temporal distance $t_2-t$ and $t-t_1$.
\eg a 2-frame video $\left[f_1, f_2\right]$ can be interpolated to 4-frame video $\left[f_1, \frac{2f_1+f_2}3, \frac{f_1+2f_2}3, f_2\right]$.
{\bf (3) Parametric interpolator} is a pre-trained interpolation network on the real video dataset.
For each video data with $L_{syn}$ frames, we evenly sample $N_{syn}$ frames and duplicate them to length $L_{syn}$. We train a TimeSformer~\cite{timesformer} model $\varphi$ on these real data and it learns to recover the original video from the duplicated one. The pretrained model $\varphi$ can be utilized for interpolation, \eg a 2-frame video $\left[f_1, f_2\right]$ can be interpolated to 4-frame video $\varphi(\left[f_1, f_1, f_2, f_2\right])$.

\begin{table}[t]
    \centering
    \resizebox{0.99\linewidth}{!}{
        \begin{tabular}{l|cccc}
        \hline
        $N_{syn}$ and $N_{real}$ &  1  &  2  &  4  & 8  \\
        \hline
        Duplication   & $17.4\pm0.3$ & $18.0\pm0.4$ & $17.3\pm0.5$ & $17.9\pm0.6$ \\
        Linear        &    -         & $15.6\pm0.1$ & $16.1\pm0.4$ & $16.3\pm0.5$ \\
        Parametric    & $17.0\pm0.6$ & $18.6\pm0.8$ & $19.2\pm1.0$ & $18.5\pm0.9$ \\
        \hline
        \end{tabular}
    } 
    \vspace{-5px}
    \caption{Comparison of different interpolation algorithm $\mathcal{I}$ with DM~\cite{DM} and ConvNet+GRU model.}
    \vspace{-9px}
    \label{tab:interpolator}
\end{table}

We compare the three interpolators in Tab.~\ref{tab:interpolator}. The simple duplication method outperforms linear interpolation. The parametric method performs the best, especially on larger frame numbers, as it encodes dataset-specific inductive bias to compensate for the loss of temporal information.

{\bf Discussion.}
With the comparison of the four dimensions of temporal compression, we have some fundamental observations that could offer direction for our further study of video distillation:
(1) Temporal compression confers significant advantages to dataset distillation and static images could encode more than little knowledge for video datasets;
(2) Segmented distillation could reduce the distillation cost, but significantly sacrifice the model performance
(3) Parametric interpolation could compensate for the loss of temporal dynamic information in the video.
\section{Methodology}
\label{sec:method}
\begin{figure}[tp]
    \centering
    \includegraphics[width=0.99\linewidth]{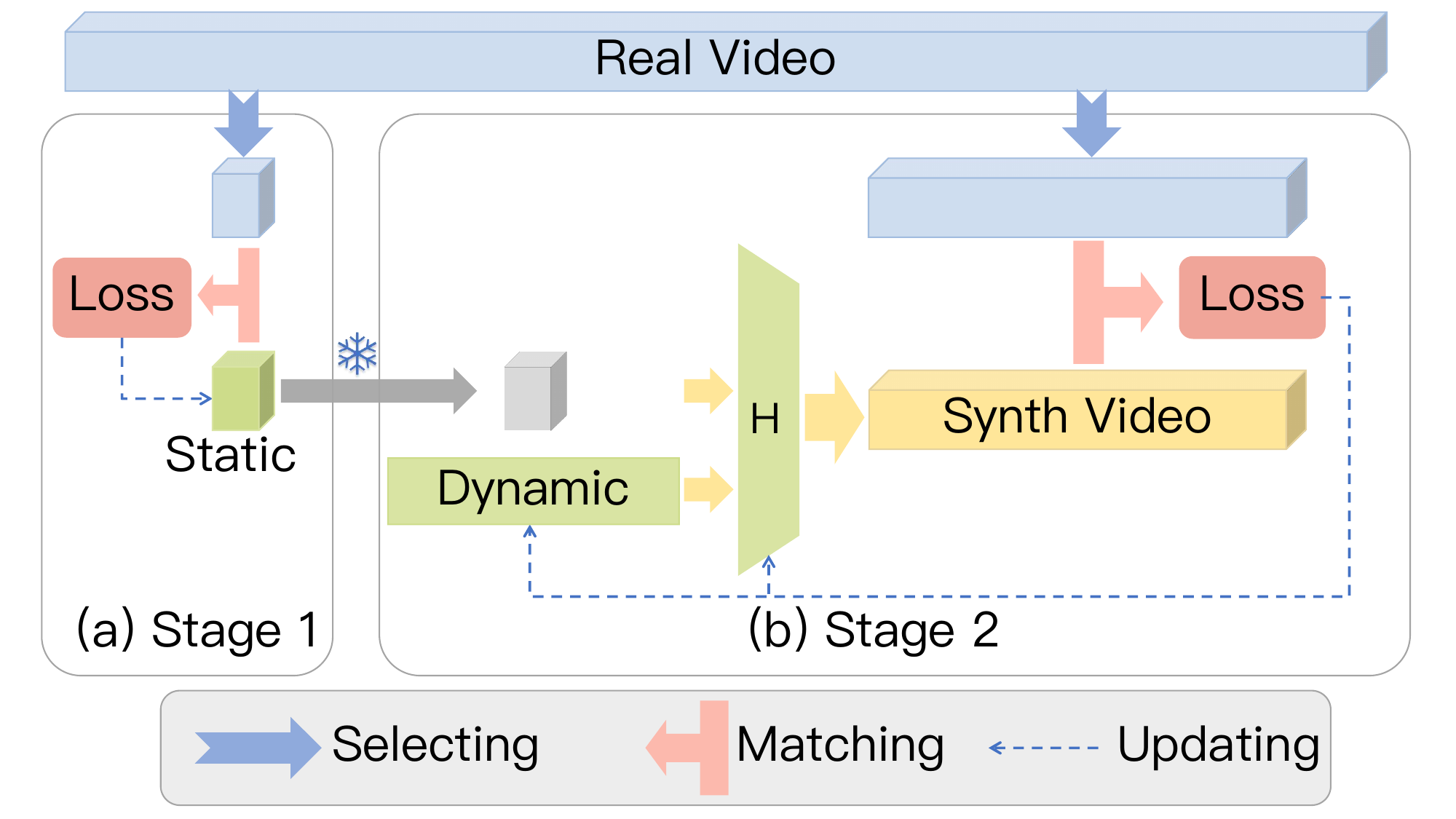}
    \vspace{-5px}
    \caption{Our two-stage method:
    Stage 1: static memory learning with image distillation on one frame per video. Stage 2: the static (frozen) and dynamic memory are combined into synthetic videos by $\mathcal{H}$, and aligned with the real data.}
    \vspace{-5px}
    \label{fig:overview}
\end{figure}

Dataset distillation is a lossy data compression process. With synthetic data as an intermediary, only part of the information in a real dataset could finally be learned by the model according to the data processing inequality.
Thus, based on the analysis in Sec.~\ref{subsec:compare-compression} and considering the trade-off between efficiency and efficacy, we propose a video dataset distillation paradigm by disentangling the static and dynamic information in videos.
We put more effort into the learning of static information with low cost (Sec.~\ref{subsec:first-stage}), and then compensate for the dynamic information (Sec.~\ref{subsec:second-stage}).
We give an overview of our method in Fig.~\ref{fig:overview}.

\subsection{Static Learning}
\label{subsec:first-stage}
The results in Sec.~\ref{subsec:compare-compression} indicate that static information in videos is more critical to the distillation task, given the limited capacity of small synthetic data.
Hence, we use a $N_{syn}=N_{real}=K=1$ setting to learn a \textbf{static memory} with only one frame. The specific distillation process involves selecting one frame randomly from each video segment to form an image dataset in each epoch. The DC~\cite{DC} method is then applied for gradient matching on a convolutional network. Since a new image dataset is created in each epoch, the ``image'' we distill has ideally observed all video frames and distilled the static memory from them.

\subsection{Dynamic Fine-tuning}
\label{subsec:second-stage}
The choice of dynamic memory can be diverse. In this paper, our dynamic memory is represented as multiple frames of single-channel images. We use a network $\mathcal{H}$ which takes static memory and dynamic memory as input and outputs video clips. At this stage, we fix the static memory and use different matching methods (performance matching, distribution matching, and parameter matching) to simultaneously update the network $\mathcal{H}$ and dynamic memory. \par
We use a concrete formula to explain our paradigm. We refer $\mathcal{A}(\mathcal{T}_{syn},\mathcal{T})$ as a matching loss of distillation where $\mathcal{T}_{syn}$ is synthetic dataset and $\mathcal{T}$ indicates origin dataset  $\{(x_i,y_i)\}_{i=1}^{|\mathcal{T}|}$, $x_i\in\mathbb{R}^{f_i\times c\times h\times w}$ , $y_i\in \{0, 1, \dots, C-1\}$. Given a frame selection method $\mathcal{B}(\mathcal{T}, N)$ selecting $N$ frames from $\mathcal{T}$ to obtain a dataset with $x_i\in\mathbb{R}^{N\times c\times h\times w}$, we summarize our paradigm in Alg.~\ref{alg1}.

\begin{algorithm}[t]
	\renewcommand{\algorithmicrequire}{\textbf{Input:}}
	\renewcommand{\algorithmicensure}{\textbf{Output:}}
	\caption{\added{Static Learning and Dynamic Fine-tuning.}} 
	\label{alg1}
        \begin{algorithmic}
        \REQUIRE Distillation matching loss $\mathcal{A}$, origin real video dataset $\mathcal{T}$, selection method $\mathcal{B}$.
        \ENSURE Static and dynamic memory $\mathcal{S}$, $\mathcal{D}$, network $\mathcal{H}$
        \STATE {\bf Stage 1: Static Learning}
        \bindent
        \STATE Initialize $\mathcal{S}$ with random frames $\mathcal{T}$.
        \FOR{$i=1$ to $M$}
        \STATE $\mathcal{T}_{s}=\mathcal{B}(\mathcal{T},1)$ \added{// Form a single-frame dataset}
        \STATE $\mathcal{S} \leftarrow \mathcal{S} - \alpha_S\nabla_S \mathcal{A}(\mathcal{S},\mathcal{T}_{s})$ \added{ // Update $\mathcal{S}$}
        \ENDFOR
        \eindent
        \STATE {\bf Stage 2: Dynamic Fine-tuning}
        \bindent
        \STATE Initialize $\mathcal{D}$ and $\mathcal{H}$ with noise
        \FOR{$i=1$ to $N$}
        \STATE $\mathcal{T}_{d}=\mathcal{B}(\mathcal{T},N_{real})$ \added{// Form a multi-frame dataset}
        \STATE $\mathcal{L}=\mathcal{A}(\mathcal{H}(\mathcal{D},\mathcal{S}),\mathcal{T}_{d})$ \added{ // Combine back to multi-frames and compute matching loss}
        \STATE $\mathcal{H} \leftarrow \mathcal{H} - \alpha_H\nabla_H \mathcal{L}$ \added{ // Update $\mathcal{H}$}
        \STATE $\mathcal{D} \leftarrow \mathcal{D} - \alpha_D\nabla_D \mathcal{L}$ \added{ // Update $\mathcal{D}$}
        \ENDFOR
        \eindent
        \end{algorithmic}
 \end{algorithm}
\section{Experiments}

\begin{table*}[ht]
    \centering
    \resizebox{\linewidth}{!}{
    \begin{subtable}{.5\linewidth}
      \centering
        \resizebox{!}{2.3cm}{
    \begin{tabular}{l|l|cc|cc} 
      \hline
      \multicolumn{2}{c|}{Dataset}       & \multicolumn{2}{c|}{MiniUCF}         &  \multicolumn{2}{c}{HMDB51} \\ 
      \multicolumn{2}{c|}{IPC}          &     1    &     5    &     1    &  5          \\
      \hline
      \multicolumn{2}{c|}{Full Dataset} & \multicolumn{2}{c|}{$57.22\pm0.14$}  &  \multicolumn{2}{c}{$28.58\pm0.69$} \\ 
      \hline
      \multirow{3}*{\makecell{Coreset\\Selection}}& Random &  $9.9\pm0.8$  &  $22.9\pm1.1$  &  $4.6\pm0.5$  &  $6.6\pm0.7$\\
 ~&Herding~\cite{herding}& $12.7\pm1.6$& $25.8\pm0.3$& $3.8\pm0.2$&$8.5\pm0.4$\\
 ~&K-Center~\cite{k-center}& $11.5\pm0.7$& $23.0\pm1.3$& $3.1\pm0.1$&$5.2\pm0.3$\\  
      \hline
      \multirow{7}*{\makecell{Dataset\\Distillation}}&DM~\cite{DM}  & $15.3\pm1.1$  &  $25.7\pm0.2$  &  $6.1\pm0.2$  &  $8.0\pm0.2$\\ 
      ~&MTT~\cite{MTT}                  & $19.0\pm0.1$  &  $28.4\pm0.7$&  $6.6\pm0.5$  &  $8.4\pm0.6$ \\ 
      ~&FRePo~\cite{FRePo}              & $20.3\pm0.5$  &  $30.2\pm1.7$&   $7.2\pm0.8$&    $9.6\pm0.7$\\ 
      \cline{2-6}
      &Static-DC& $13.7\pm1.1$  &  $24.7\pm0.5$  &  $5.1\pm0.9$  &    $7.8\pm0.4$ \\  
      \cline{2-6}
      & DM~\cite{DM}+Ours    & $17.5\pm0.1$  &  $27.2\pm0.4$&  $6.0\pm0.4$&  $8.2\pm0.1$\\
      & MTT~\cite{MTT}+Ours                       & \boldmath$23.3\pm0.6$&  $28.3\pm0.0$  &  $6.5\pm0.1$  &  $8.9\pm0.6$ \\
      & FRePo~\cite{FRePo}+Ours                     & $22.0\pm1.0$&  \boldmath$31.2\pm0.7$&               \boldmath$8.6\pm0.5$&     \boldmath$10.3\pm0.6$\\
      \hline
    \end{tabular}}
    \caption{Accuracy}
    \end{subtable} 

\hspace{14mm}

    \begin{subtable}{.49\linewidth}
      \centering
        \resizebox{!}{2.3cm}{
    \begin{tabular}{l|cc|cc} 
      \hline
        Dataset&\multicolumn{2}{c|}{MiniUCF}         &  \multicolumn{2}{c}{HMDB51} \\ 
            IPC&1    &     5    &     1    &  5          \\
         \hline
         Full Dataset&\multicolumn{2}{c|}{\textcolor{red}{9.81 GB}}  &  \multicolumn{2}{c}{\textcolor{red}{4.93 GB}} \\ 
      \hline 
          Random &\multirow{3}*{\textcolor{Goldenrod}{115 MB}}&  \multirow{3}*{\textcolor{Goldenrod}{586 MB}}&  \multirow{3}*{\textcolor{Goldenrod}{115 MB}}&  \multirow{3}*{\textcolor{Goldenrod}{586 MB}}\\
   Herding~\cite{herding}&& & &\\
   K-Center~\cite{k-center}&& & &\\  
      \hline
        DM~\cite{DM}&\multirow{3}*{\textcolor{Goldenrod}{115 MB}}&  \multirow{3}*{\textcolor{Goldenrod}{586 MB}}&  \multirow{3}*{\textcolor{Goldenrod}{115 MB}}&  \multirow{3}*{\textcolor{Goldenrod}{586 MB}}\\ 
        MTT~\cite{MTT}&&  &  &  \\ 
        FRePo~\cite{FRePo}&&  &   &    \\ 
      \hline
        Static-DC&\textcolor{YellowGreen}{8 MB}&  \textcolor{YellowGreen}{36 MB}&  \textcolor{YellowGreen}{8 MB}&    \textcolor{YellowGreen}{36 MB}\\ 
      \hline
        DM~\cite{DM}+Ours           &\textcolor{YellowGreen}{94 MB}&  \textcolor{YellowGreen}{455 MB}&  \textcolor{YellowGreen}{94 MB}&  \textcolor{YellowGreen}{455 MB}\\
        MTT~\cite{MTT}+Ours                              &\textcolor{YellowGreen}{94 MB}&  \textcolor{YellowGreen}{455 MB}&  \textcolor{YellowGreen}{94 MB}&  \textcolor{YellowGreen}{455 MB}  \\
        FRePo~\cite{FRePo}+Ours                            &\textcolor{YellowGreen}{48 MB}&  \textcolor{YellowGreen}{228 MB}&  \textcolor{YellowGreen}{48 MB}&     \textcolor{YellowGreen}{228 MB}\\
      \hline
    \end{tabular}}
    \caption{Storage}
    \end{subtable} 
    }
    \vspace{-15px}
    \caption{
    Results of baselines and our method on small-scale datasets. Top-1 test accuracies (\%) and memory storage budget (MB or GB) are reported.
    \added{Storage represents the \textbf{total size of tensors}, assuming the data is stored as floats.} Our method uses no more than \textbf{42\%} storage compared with the naively adapted method for FRePo, while 82\% for DM and MTT. 
    We use the storage of coreset selection methods as a reference and color-code the \textcolor{red}{high}, \textcolor{Goldenrod}{comparable}, and \textcolor{YellowGreen}{low} storage. 
    IPC: Instance(s) Per Class.} 
    \vspace{-5px}
    \label{tab:small-scale}
\end{table*}

\begin{table}[htp]
    \centering
    \resizebox{0.8\linewidth}{!}{
    \begin{tabular}{l|cc|cc} 
      \hline
        Dataset&  \multicolumn{2}{c|}{Kinetics-400} &\multicolumn{2}{c}{\added{SSv2}}\\
        IPC&  1&  5 & 1 & 5\\ 
        \hline 
        Full Dataset &  \multicolumn{2}{c|}{$34.6\pm0.5$} &  \multicolumn{2}{c}{$29.0\pm0.6$}\\ 
        \hline
        Random&  $3.0\pm0.1$&  $5.6\pm0.0$&$3.3\pm0.1$&$3.9\pm0.1$ \\
        DM~\cite{DM}& $6.3\pm0.0$&$9.1\pm0.9$&$3.6\pm0.0$&$4.1\pm0.0$\\
        MTT~\cite{MTT}& $3.8\pm0.2$&$9.1\pm0.3$&$3.9\pm0.1$&$6.3\pm0.3$\\
        \hline
        Static-DC& $4.6\pm0.2$&$6.6\pm0.2$&$3.9\pm0.1$ &$4.1\pm0.0$\\\hline
        DM\cite{DM}+Ours& $6.3\pm0.2$&$7.0\pm0.1$&$4.0\pm0.1$&$3.8\pm0.1$\\ 
        MTT\cite{MTT}+Ours & $6.3\pm0.1$& \boldmath$11.5\pm0.5$&\boldmath$5.5\pm0.1$ &\boldmath$8.3\pm0.2$\\
      \hline
    \end{tabular}}
    \vspace{-5px}
    \caption{Top-5 accuracy on Kinetics-400~\cite{kinetics} \added{and SSv2~\cite{ssv2}}.}
    \label{tab:large-scale}
    \vspace{-7px}
\end{table}

\subsection{Datasets and Metrics}

We adopt small video datasets UCF101~\cite{ucf101} and HMDB51~\cite{hmdb51}, and large-scale Kinetics~\cite{kinetics} 
and \added{Something-Something V2~\cite{ssv2}} in our experiments.
UCF101~\cite{ucf101} consists of 13,320 video clips in 101 action categories while HMDB51~\cite{hmdb51} consists of 6849 video clips in 51 action categories. Kinetics~\cite{kinetics} is a collection of video clips that cover 400/600/700 human action classes \added{while SSv2~\cite{ssv2} covers 174 motion-heavy classes}.
To evaluate the distillation algorithms on more diversified data scales, and considering the efficiency of experiments and the clarity of model comparisons, following the scale of pioneering work on image distillation \cite{DD}, we build a miniaturized version of UCF101, named \textbf{MiniUCF}, including 50 most common classes from the UCF101 dataset. This miniaturization enables rapid iterations of our method and facilitates observing relatively significant changes in performance. 
We report the top-1 classification accuracy for MiniUCF and HMDB51, and the top-5 classification accuracy for Kinetics400 and \added{SSv2}.

\subsection{Baselines}
The baseline methods involve:
(1) coreset selection methods (random selection, Herding~\cite{herding} and K-Center~\cite{k-center}) following the implementation for image distillation in DC~\cite{DC}.
(2) direct adaptation of the common image distillation methods (DM~\cite{DM}, MTT~\cite{MTT}, FRePo~\cite{FRePo}) to the video distillation task.
(3) image distillation method (DC~\cite{DC}) with frame duplication for a ``boring videos'' proposed by us, namely ``Static-DC''.

\subsection{Implementation Details}
{\bf Data.}
For MiniUCF and HMDB51, the videos are sampled to 16 frames with sampling interval 4 dynamically, $i.e.$. the frames indices vary in different epochs. 
Following the setup in C3D~\cite{c3d}, each of these frames is cropped and resized to 112x112.
For Kinetics-400 \added{and Something-Something V2}, the videos are sampled to 8 frames before the distillation, and the frames are cropped to 64x64.
We only use horizontal flipping with a 50\% probability for data augment. 

{\bf Static Learning.}
We use DC~\cite{DC} to distill static memory with random real frame initialization. We utilize a 4-layer 2D convolutional neural network for distillation (ConvNetD4)
and perform an early stop in the distillation training when the loss converges. 
Interestingly, our experiments indicate that static memory is not necessarily trained to full convergence, as dynamic memory compensates for it. 

{\bf Dynamic Finetuning.}
In dynamic fine-tuning, we adopt distillation methods in various types, including DM~\cite{DM}, MTT~\cite{MTT}, and FRePo~\cite{FRePo} to evaluate the broad applicability of our paradigm. Dynamic memory is initialized with random noise. We use a small 3D CNN (referred to as MiniC3D) for distillation. 
The $\mathcal{H}$ network used for combining static and dynamic memory is also a MiniC3D.
For more details, please refer to the supplementary. 

{\bf Fair Comparison}
\label{sec: fair comparison}
between the baseline and our method. We rigorously ensure that the total storage space for static memory, dynamic memory, and the $\mathcal{H}$ network is smaller than the corresponding IPC (Instance Per Class). Specifically, on DM and MTT, we use no more than 82\% of the storage space corresponding to the baseline, which amounts to 2 static memory with 2 dynamic memory for every instance. On FRePo, we use no more than 42\% of the storage space corresponding to the baseline, which means 1 static memory with 1 dynamic memory for every instance.

{\bf Evaluation of Distilled Dataset.}
Naturally, we evaluate how well our synthetic data performs on architectures used to distill it. When evaluating data distilled by FRePo, the results should be considered as a reference only due to the label learning conducted by the method itself and the use of a different optimizer.
We also evaluate how well our synthetic data performs on different architectures from the one used to distill it on the MiniUCF, 1 instance per class task.

{\bf Hyper-Parameters.}
Considering the numerous parameters involved in the experiments, we detail the parameter settings for all experiments in the supplementary.

\subsection{Results}
We show the results of our small-scale experiments in Tab.~\ref{tab:small-scale} and large-scale in Tab.~\ref{tab:large-scale}. The full dataset indicates the accuracy of the network trained on the full dataset. 

{\bf Comparison to Coreset Method.}
Following image distillation, we compare our method with coreset selection methods on MiniUCF and HMDB51 in Tab.~\ref{tab:small-scale}. In the majority of cases, our approach outperforms the coreset selection. 
Regarding coreset methods, we also observe: 
(1) On average, the herding method proves to be the most effective coreset approach. 
(2) With an increase in IPC, the performance of herding exhibits a notable improvement. 
These conclusions align with earlier experiments~\cite{DC} in image distillation, lending credibility to our findings.

{\bf Comparison to Other Methods.}
We compare our \added{final} method with other methods we proposed in Fig.~\ref{fig:teaser}. Among the three methods we proposed, Static-DC \added{(Fig.~\ref{fig:teaser}(b))} exhibits the poorest performance.
Compared to both the coreset method and Static-DC, the naively adapted method \added{(Fig.~\ref{fig:teaser}(a))} shows a significant improvement. This underscores the applicability of existing image distillation techniques to videos and the effects of dynamic information in video understanding. 
In comparison, our final method \added{(corresponding to Fig.~\ref{fig:teaser}(c))} achieves a remarkable superiority over all other methods through static and dynamic disentangle learning. In most cases, our method could enhance the current distillation methods while requiring less storage.
However, the performance of our method on Kinetics-400 is not as strong as on the other two smaller datasets. 
This is because Kinetics-400 itself has a much larger number of categories and samples compared to the other two smaller datasets, which increases the difficulty and cost of distillation. Since our $\mathcal{H}$ network is shared, having 400 categories sharing one $\mathcal{H}$ network in Kinetics-400 might cause interference during learning. Allocating different $\mathcal{H}$ networks to different categories or using a more complex $\mathcal{H}$ network could potentially offer better improvement in our method. However, this approach would impose a greater training burden and significantly reduce the practical value of distillation, so we do not conduct larger-scale experiments. 

{\bf Cross Architecture Generalization.}
We also show the result of cross-architecture generalization in Tab.~\ref{tab:cross-arch}. The experimental results indicate that the data obtained by static-dynamic disentanglement performs much better on other networks compared to the naively adapted method. 
\begin{table}[t]
    \centering
    \resizebox{0.85\linewidth}{!}{
    \begin{tabular}{l|ccc} 
    \hline
         &  \multicolumn{3}{c}{Evaluation Model}\\
        & ConvNet3D& CNN+GRU& CNN+LSTM\\ \hline
        Random & $9.9\pm0.8$& $6.2\pm0.8$&$6.5\pm0.3$ \\ \hline
        DM~\cite{DM}& $15.3\pm1.1$  & $9.9\pm0.7$& $9.2\pm0.3$\\
         DM\cite{DM}+Ours& \boldmath$17.5\pm0.1$  & \boldmath$12.0\pm0.7$& \boldmath$10.3\pm0.2$\\
         \hline
         MTT~\cite{MTT}&  $19.0\pm0.1$&  $8.4\pm0.5$&  $7.3\pm0.4$\\ 
         MTT\cite{MTT}+Ours &  \boldmath$23.3\pm0.6$&  \boldmath$14.8\pm0.1$&  \boldmath$13.4\pm0.2$\\
    \hline
    \end{tabular}}
    \vspace{-5px}
    \caption{Cross-architecture generalization for MiniUCF IPC=1. CNN+GRU and CNN+LSTM are detailed in the supplementary.}
    \vspace{-10px}
    \label{tab:cross-arch}
\end{table}

\subsection{Ablation Study}
{\bf Ratios of Static and Dynamic.}
We report results for MiniUCF 1 Instance per class task with different static and dynamic memory ratios in Tab.~\ref{tab:ablation}. 
We can find that increasing the quantity of static and dynamic memory both improves the scores. However, as their numbers increase, the time and GPU memory required for Static Learning and the training convergence time for Dynamic Fine-tuning under the same computational power will also increase (Tab.~\ref{tab:ablation}). Considering training efficiency and effectiveness while ensuring the storage does not exceed the corresponding IPC, a balanced choice is to use 2 static with 2 dynamics for every instance. 

\begin{figure}
    \centering
    \includegraphics[width=0.45\textwidth]{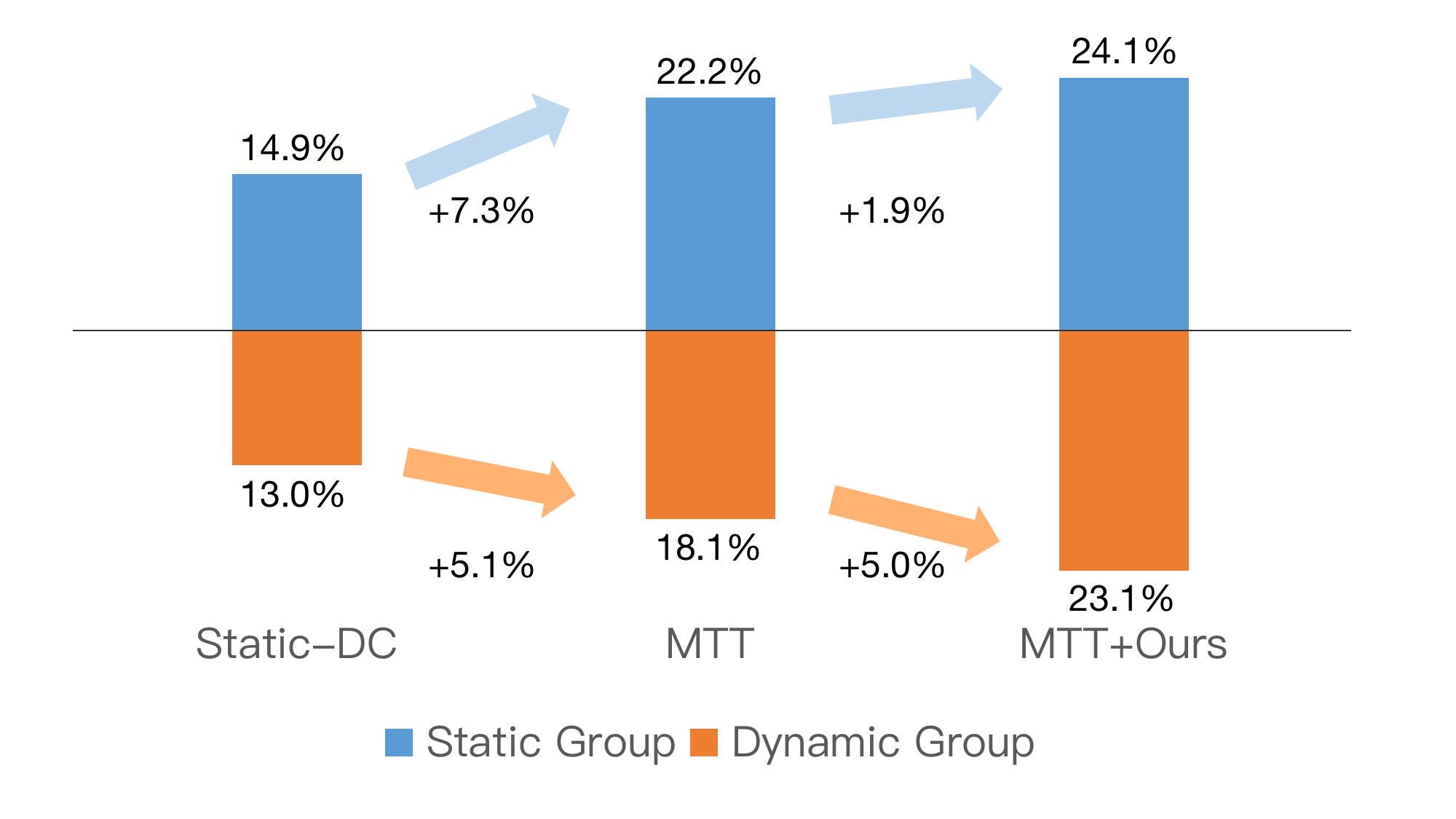}
    \vspace{-15px}
    \caption{Test accuracies of static and dynamic group on network trained with distilled data.}
    \label{fig:s_d}
    \vspace{-8px}
\end{figure}

\begin{table}[t]
    \centering
    \resizebox{0.95\linewidth}{!}{
    \begin{tabular}{cc|cccc}
    \hline
         SPC & DPC & Acc (\%)& Storage & S1 Time& S1 GPU Memory\\ \hline
         \multirow{4}*{1}& 0&  $13.7\pm1.1$ & 8 MB& \multirow{4}*{68~s/iter}&\multirow{4}*{4,651 MiB}\\
            ~& 1&$ 17.5\pm0.5$ &48 MB&~&~\\
         ~& 2&  $19.6\pm1.2$&79 MB&~&~\\
            ~& 3&$20.6\pm0.2$&111 MB&~&~\\ \hline
         \multirow{4}*{2}& 0 &  $20.4\pm0.5$&14 MB&\multirow{4}*{156~s/iter}&\multirow{4}*{6,579 MiB}\\
            ~& 1 & $22.3\pm0.0$&54 MB&~&~\\
         ~&  2& $23.3\pm0.6$ & 94 MB&~&~\\
            ~& 3& -&\textcolor{red}{117 MB}&~&~\\ 
    \hline
    \end{tabular}}
    \vspace{-5px}
    \caption{Results and stage 1 cost for MiniUCF IPC=1 with different ratios of static and dynamic memory. SPC: static memory per class. DPC: dynamic memory per class. For MiniUCF IPC=1, the storage for naive methods is \textcolor{red}{115 MB}. We ensure that the storage used by our method is less than 115 MB. S1 Time: The time required for each iteration (s/iter) at Stage 1 (static learning stage). S1 GPU Memory: The GPU memory (MiB) required at Stage 1. }
    \label{tab:ablation}
    \vspace{-5px}
\end{table}
{\bf Impact of Video Dynamics on Distillation.} Actions in the video exhibit varying degrees of dynamics. To explore the impact of video dynamics, we categorized all classes of MiniUCF into two groups (relatively static, highly dynamic) based on their level of dynamics, calculated by the average Hamming distance between inter-frame features for each class, and compared their test accuracy on networks trained with distilled data (Fig.~\ref{fig:s_d}). We observe that (1) data distilled by the Static-DC method is more sensitive for static classes, which aligns with our expectations as this method generates data that lacks dynamic information, akin to a ``boring video''. (2) The naively adapted MTT, in comparison, can distill more useful information but still shows higher scores for static classes than dynamic classes. (3) MTT+Ours, however, demonstrates better distillation of dynamic information compared to the previous methods and exhibits significant improvements on dynamic classes.

\subsection{Visualization}
To observe the temporal changes in the distilled videos, we sampled frames from the videos obtained using different methods and visualized their \textit{inter-frame differences}. We show two examples in Fig.~\ref{fig:static} and more in the supplementary. Although visually abstract, we can still conclude that the distilled videos indeed exhibit temporal variations.

\begin{figure}
    \centering
    \includegraphics[width=1\linewidth]{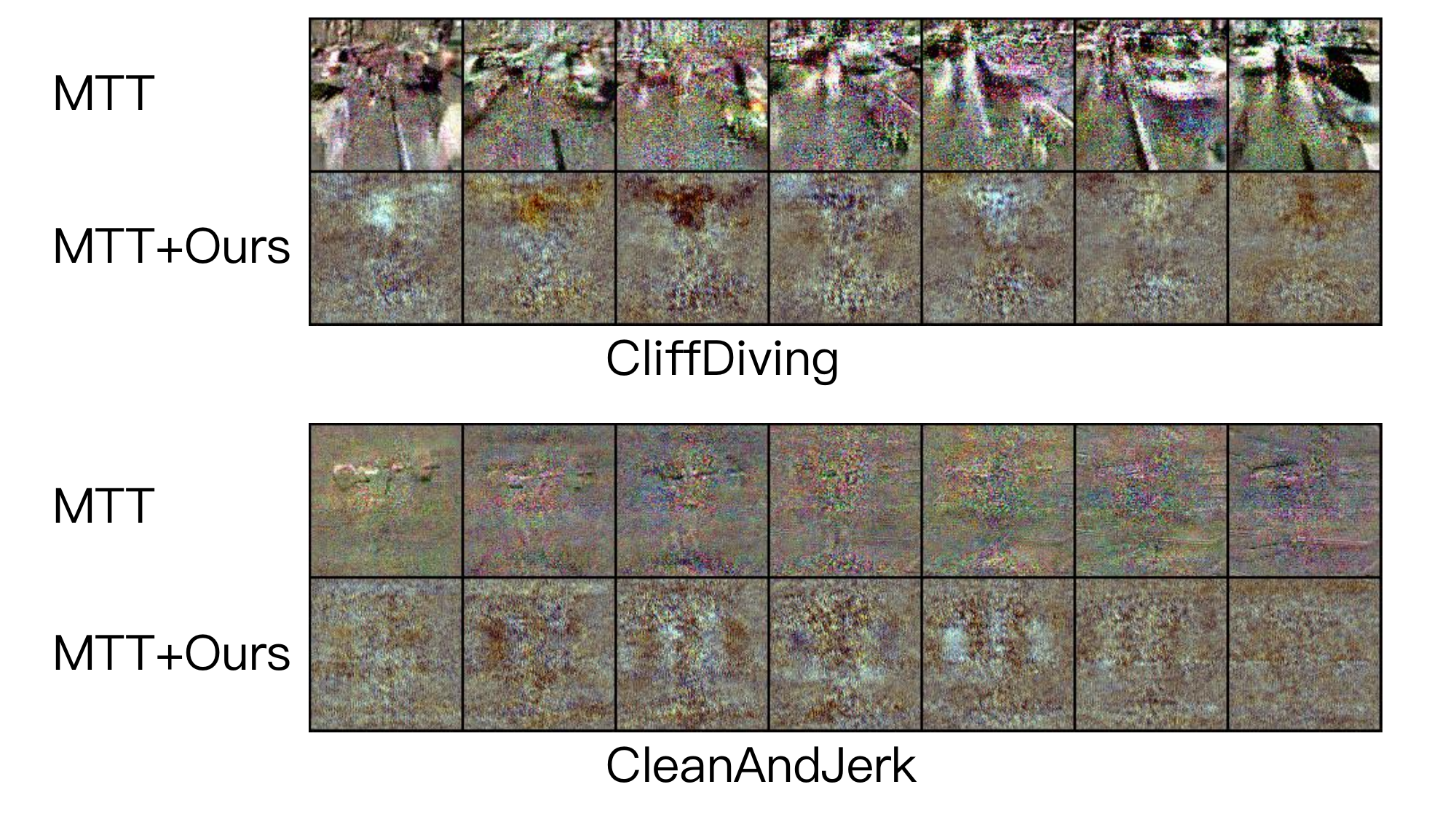}
    \vspace{-15px}
    \caption{Visualized inter-frame differences of videos distilled by MTT and MTT+Ours for MiniUCF IPC=1.}
    \label{fig:static}
    \vspace{-7px}
\end{figure}
\section{Conclusion}
In this work, we provide the first systematic study of video distillation.
We propose a taxonomy to categorize methods based on key factors including the number of frames and segment length. 
With extensive experiments, we revealed that more frames provide marginal gains at greatly increased costs. 
Then, we proposed our method that disentangles static and dynamic information and achieves SOTA with efficient storage. 
We believe our paradigm will pave a new way for video distillation.
\section*{Acknowledgments} 
This work is supported in part by the
National Natural Science Foundation of China under Grants 62306175, 
National Key Research and Development Project of China (No.2022ZD0160102, No.2021ZD0110704), Shanghai Artificial Intelligence Laboratory, XPLORER PRIZE grants.
{
    \small
    \bibliographystyle{ieeenat_fullname}
    \bibliography{main}

\begin{thebibliography}{32}
\providecommand{\natexlab}[1]{#1}
\providecommand{\url}[1]{\texttt{#1}}
\expandafter\ifx\csname urlstyle\endcsname\relax
  \providecommand{\doi}[1]{doi: #1}\else
  \providecommand{\doi}{doi: \begingroup \urlstyle{rm}\Url}\fi

\bibitem[Baker et~al.(2011)Baker, Scharstein, Lewis, Roth, Black, and Szeliski]{opflow}
Simon Baker, Daniel Scharstein, JP Lewis, Stefan Roth, Michael~J Black, and Richard Szeliski.
\newblock A database and evaluation methodology for optical flow.
\newblock \emph{IJCV}, 92:\penalty0 1--31, 2011.

\bibitem[Bertasius et~al.(2021)Bertasius, Wang, and Torresani]{timesformer}
Gedas Bertasius, Heng Wang, and Lorenzo Torresani.
\newblock Is space-time attention all you need for video understanding?
\newblock In \emph{ICML}, page~4, 2021.

\bibitem[Carreira and Zisserman(2017)]{kinetics}
João Carreira and Andrew Zisserman.
\newblock Quo vadis, action recognition? a new model and the kinetics dataset.
\newblock In \emph{CVPR}, pages 4724--4733, 2017.

\bibitem[Cazenavette et~al.(2022)Cazenavette, Wang, Torralba, Efros, and Zhu]{MTT}
George Cazenavette, Tongzhou Wang, Antonio Torralba, Alexei~A Efros, and Jun-Yan Zhu.
\newblock Dataset distillation by matching training trajectories.
\newblock In \emph{CVPR}, 2022.

\bibitem[Cho et~al.(2014)Cho, Van~Merri{\"e}nboer, Bahdanau, and Bengio]{GRU}
Kyunghyun Cho, Bart Van~Merri{\"e}nboer, Dzmitry Bahdanau, and Yoshua Bengio.
\newblock On the properties of neural machine translation: Encoder-decoder approaches.
\newblock \emph{arXiv preprint arXiv:1409.1259}, 2014.

\bibitem[Cui et~al.(2023)Cui, Wang, Si, and Hsieh]{tesla}
Justin Cui, Ruochen Wang, Si Si, and Cho-Jui Hsieh.
\newblock Scaling up dataset distillation to imagenet-1k with constant memory.
\newblock In \emph{ICML}, pages 6565--6590, 2023.

\bibitem[Deng and Russakovsky(2022)]{linba}
Zhiwei Deng and Olga Russakovsky.
\newblock Remember the past: Distilling datasets into addressable memories for neural networks.
\newblock In \emph{NeurIPS}, 2022.

\bibitem[Goyal et~al.(2017)Goyal, Kahou, Michalski, Materzyńska, Westphal, Kim, Haenel, Fruend, Yianilos, Mueller-Freitag, Hoppe, Thurau, Bax, and Memisevic]{ssv2}
Raghav Goyal, Samira~Ebrahimi Kahou, Vincent Michalski, Joanna Materzyńska, Susanne Westphal, Heuna Kim, Valentin Haenel, Ingo Fruend, Peter Yianilos, Moritz Mueller-Freitag, Florian Hoppe, Christian Thurau, Ingo Bax, and Roland Memisevic.
\newblock The "something something" video database for learning and evaluating visual common sense, 2017.

\bibitem[Huang et~al.(2018)Huang, Ramanathan, Mahajan, Torresani, Paluri, Fei-Fei, and Niebles]{wmvv}
De-An Huang, Vignesh Ramanathan, Dhruv Mahajan, Lorenzo Torresani, Manohar Paluri, Li Fei-Fei, and Juan~Carlos Niebles.
\newblock What makes a video a video: Analyzing temporal information in video understanding models and datasets.
\newblock In \emph{CVPR}, pages 7366--7375, 2018.

\bibitem[Kim et~al.(2022)Kim, Kim, Oh, Yun, Song, Jeong, Ha, and Song]{IDC}
Jang-Hyun Kim, Jinuk Kim, Seong~Joon Oh, Sangdoo Yun, Hwanjun Song, Joonhyun Jeong, Jung-Woo Ha, and Hyun~Oh Song.
\newblock Dataset condensation via efficient synthetic-data parameterization.
\newblock In \emph{ICML}, 2022.

\bibitem[Kuehne et~al.(2011)Kuehne, Jhuang, Garrote, Poggio, and Serre]{hmdb51}
H. Kuehne, H. Jhuang, E. Garrote, T. Poggio, and T. Serre.
\newblock Hmdb: A large video database for human motion recognition.
\newblock In \emph{ICCV}, pages 2556--2563, 2011.

\bibitem[Li and Li(2021)]{li2021data-TEXT}
Yongqi Li and Wenjie Li.
\newblock Data distillation for text classification.
\newblock \emph{arXiv preprint arXiv:2104.08448}, 2021.

\bibitem[Liu et~al.(2022)Liu, Wang, Yang, Ye, and Wang]{haba}
Songhua Liu, Kai Wang, Xingyi Yang, Jingwen Ye, and Xinchao Wang.
\newblock Dataset distillation via factorization.
\newblock In \emph{NeurIPS}, 2022.

\bibitem[Liu et~al.(2021)Liu, Pintea, Nejadasl, Booij, and van Gemert]{nflb}
Xin Liu, Silvia~L. Pintea, Fatemeh~Karimi Nejadasl, Olaf Booij, and Jan~C. van Gemert.
\newblock No frame left behind: Full video action recognition.
\newblock In \emph{CVPR}, pages 14892--14901, 2021.

\bibitem[Loo et~al.(2022)Loo, Hasani, Amini, and Rus]{RFAD}
Noel Loo, Ramin Hasani, Alexander Amini, and Daniela Rus.
\newblock Efficient dataset distillation using random feature approximation.
\newblock In \emph{NeurIPS}, 2022.

\bibitem[Nguyen et~al.(2020)Nguyen, Chen, and Lee]{KIP}
Timothy Nguyen, Zhourong Chen, and Jaehoon Lee.
\newblock Dataset meta-learning from kernel ridge-regression.
\newblock \emph{arXiv preprint arXiv:2011.00050}, 2020.

\bibitem[Nguyen et~al.(2021)Nguyen, Novak, Xiao, and Lee]{KIP2}
Timothy Nguyen, Roman Novak, Lechao Xiao, and Jaehoon Lee.
\newblock Dataset distillation with infinitely wide convolutional networks.
\newblock In \emph{NeurIPS}, 2021.

\bibitem[Sener and Savarese(2018)]{k-center}
Ozan Sener and Silvio Savarese.
\newblock Active learning for convolutional neural networks: A core-set approach.
\newblock In \emph{ICLR}, 2018.

\bibitem[Soomro et~al.(2012)Soomro, Zamir, and Shah]{ucf101}
Khurram Soomro, Amir~Roshan Zamir, and Mubarak Shah.
\newblock {UCF101:} {A} dataset of 101 human actions classes from videos in the wild.
\newblock \emph{CoRR}, abs/1212.0402, 2012.

\bibitem[Tong et~al.(2022)Tong, Song, Wang, and Wang]{videovae}
Zhan Tong, Yibing Song, Jue Wang, and Limin Wang.
\newblock Videomae: Masked autoencoders are data-efficient learners for self-supervised video pre-training.
\newblock In \emph{NeurIPS}, pages 10078--10093. Curran Associates, Inc., 2022.

\bibitem[Tran et~al.(2015)Tran, Bourdev, Fergus, Torresani, and Paluri]{c3d}
Du Tran, Lubomir Bourdev, Rob Fergus, Lorenzo Torresani, and Manohar Paluri.
\newblock Learning spatiotemporal features with 3d convolutional networks.
\newblock In \emph{ICCV}, pages 4489--4497, 2015.

\bibitem[Tran et~al.(2018)Tran, Wang, Torresani, Ray, LeCun, and Paluri]{r2+1d}
Du Tran, Heng Wang, Lorenzo Torresani, Jamie Ray, Yann LeCun, and Manohar Paluri.
\newblock A closer look at spatiotemporal convolutions for action recognition.
\newblock In \emph{CVPR}, pages 6450--6459, 2018.

\bibitem[Wang et~al.(2022)Wang, Zhao, Peng, Zhu, Yang, Wang, Huang, Bilen, Wang, and You]{CAFE}
Kai Wang, Bo Zhao, Xiangyu Peng, Zheng Zhu, Shuo Yang, Shuo Wang, Guan Huang, Hakan Bilen, Xinchao Wang, and Yang You.
\newblock Cafe: Learning to condense dataset by aligning features.
\newblock In \emph{CVPR}, 2022.

\bibitem[Wang et~al.(2021)Wang, Xing, and Liu]{actionclip}
Mengmeng Wang, Jiazheng Xing, and Yong Liu.
\newblock Actionclip: {A} new paradigm for video action recognition.
\newblock \emph{CoRR}, abs/2109.08472, 2021.

\bibitem[Wang et~al.(2018)Wang, Zhu, Torralba, and Efros]{DD}
Tongzhou Wang, Jun-Yan Zhu, Antonio Torralba, and Alexei~A Efros.
\newblock Dataset distillation.
\newblock \emph{arXiv preprint arXiv:1811.10959}, 2018.

\bibitem[Welling(2009)]{herding}
Max Welling.
\newblock Herding dynamical weights to learn.
\newblock In \emph{ICML}, 2009.

\bibitem[Wu et~al.(2023)Wu, Deng, and Russakovsky]{wu2023multimodal-CLIP}
Xindi Wu, Zhiwei Deng, and Olga Russakovsky.
\newblock Multimodal dataset distillation for image-text retrieval.
\newblock \emph{arXiv preprint arXiv:2308.07545}, 2023.

\bibitem[Xu et~al.(2023)Xu, Chen, Pan, Chen, Das, Yang, and Tong]{xu2023kernel-GRAPH}
Zhe Xu, Yuzhong Chen, Menghai Pan, Huiyuan Chen, Mahashweta Das, Hao Yang, and Hanghang Tong.
\newblock Kernel ridge regression-based graph dataset distillation.
\newblock In \emph{Proceedings of the 29th ACM SIGKDD Conference on Knowledge Discovery and Data Mining}, pages 2850--2861, 2023.

\bibitem[Zhao and Bilen(2021)]{DSA}
Bo Zhao and Hakan Bilen.
\newblock Dataset condensation with differentiable siamese augmentation.
\newblock In \emph{ICML}, 2021.

\bibitem[Zhao and Bilen(2023)]{DM}
Bo Zhao and Hakan Bilen.
\newblock Dataset condensation with distribution matching.
\newblock In \emph{WACV}, 2023.

\bibitem[Zhao et~al.(2020)Zhao, Mopuri, and Bilen]{DC}
Bo Zhao, Konda~Reddy Mopuri, and Hakan Bilen.
\newblock Dataset condensation with gradient matching.
\newblock \emph{arXiv preprint arXiv:2006.05929}, 2020.

\bibitem[Zhou et~al.(2022)Zhou, Nezhadarya, and Ba]{FRePo}
Yongchao Zhou, Ehsan Nezhadarya, and Jimmy Ba.
\newblock Dataset distillation using neural feature regression.
\newblock \emph{arXiv preprint arXiv:2206.00719}, 2022.

\end{thebibliography}
}
\clearpage

\setcounter{page}{1}
\maketitlesupplementary
\startcontents
\printcontents{}{1}{}

\section{Selection of MiniUCF}
We train and test on split 1 of UCF101 on MiniC3D and select the top 50 classes based on accuracy. MiniUCF can reach an accuracy of \textbf{57.2\% }on MiniC3D. We provide the 50 categories we have selected in Suppl.~Tab.~\ref{tab:miniucf}.
\begin{table}[h]
    \centering
    \resizebox{0.95\linewidth}{!}{
    \begin{tabular}{lll}
    \hline
         \textcolor{blue}{ApplyEyeMakeup}       &  \textcolor{orange}{BalanceBeam}      &  \textcolor{orange}{BandMarching} \\
         \textcolor{blue}{BaseballPitch}        & \textcolor{blue}{Basketball}          &  \textcolor{orange}{BasketballDunk} \\
         \textcolor{orange}{Biking}             &  \textcolor{blue}{ Billiards}         &  \textcolor{blue}{BlowingCandles} \\
         \textcolor{blue}{Bowling}              & \textcolor{orange}{BreastStroke}      &  \textcolor{blue}{CleanAndJerk}   \\
         \textcolor{orange}{CliffDiving}        &  \textcolor{blue}{CricketShot}        &  \textcolor{blue}{ Diving}  \\
         \textcolor{orange}{FloorGymnastics}    &  \textcolor{orange} {FrisbeeCatch}    &  \textcolor{blue}{GolfSwing}     \\ 
         \textcolor{blue}{HammerThrow}          & \textcolor{orange}{HighJump}          & \textcolor{orange}{HorseRace}         \\
         \textcolor{orange}{HorseRiding}        &  \textcolor{blue}{HulaHoop}           &  \textcolor{orange}{IceDancing} \\
         \textcolor{blue}{ JumpingJack}         & \textcolor{orange}{Knitting}          &  \textcolor{orange}{MilitaryParade}   \\
         \textcolor{orange} {Mixing}            &  \textcolor{blue}{ParallelBars}       & \textcolor{blue}{PlayingPiano}   \\
         \textcolor{blue}{PlayingViolin}        &  \textcolor{orange}{PoleVault}        &  \textcolor{blue}{PommelHorse} \\
         \textcolor{orange}{Punch}              & \textcolor{orange}{Rafting}           & \textcolor{blue}{Rowing}            \\
         \textcolor{orange} {SkateBoarding}     &  \textcolor{orange}{Skiing}           &  \textcolor{orange}{Skijet}   \\
         \textcolor{orange}{SkyDiving}          & \textcolor{blue}{SoccerPenalty}       &  \textcolor{blue}{ StillRings}     \\
         \textcolor{blue}{SumoWrestling}        &  \textcolor{blue}{Surfing}            & \textcolor{orange}{Swing}     \\
         \textcolor{blue}{TennisSwing}          &  \textcolor{orange}{TrampolineJumping}&  \textcolor{orange}{UnevenBars}  \\
         \textcolor{blue}{ VolleyballSpiking}   & \textcolor{blue}{WritingOnBoard}   \\
         \hline
    \end{tabular}}
    \caption{Action classes in the adopted. We highlight the static group with \textcolor{blue}{blue} and the dynamic group with \textcolor{orange}{orange}.}
    \label{tab:miniucf}
\end{table}

\section{\added{Ablation Study of Uniformity}}

\added{We conduct experiments to justify the necessity of uniformity in Supp.~Tab.~\ref{tab:exp-uniform}. Eight real frames are split evenly or non-evenly into 4 segments and aligned to 4 synthetic frames, among which uniform segmentation achieves the best performance.}

\begin{table}[t]
    \centering
    \resizebox{0.97\linewidth}{!}{
    \begin{tabular}{c|c||c|c} \hline
     Segment Sizes & Acc. & Segment Sizes & Acc.\\ 
        \hline
        1,1,5,1 & 15.8$\pm$0.2 & 1,1,1,5 & 15.7$\pm$0.5\\
        1,1,2,4 & 14.1$\pm$0.2 & 1,4,1,2 & 15.1$\pm$0.2\\
        3,1,2,2 & 17.6$\pm$0.7 & 1,2,2,3 & 16.8$\pm$0.1 \\
        \hline
        2,2,2,2 (uniform) & \textbf{17.6$\pm$0.2} \\
        \hline
    \end{tabular}}
    \caption{\added{Results on UCF101 with different uniormity: $N_{real}=8$ are split unevenly to $N_{syn}=4$ segments}.}
    \label{tab:exp-uniform}
\end{table}

\begin{table}[t]
    \centering
    \resizebox{0.65\linewidth}{!}{
    \begin{tabular}{cc|ccc} \hline
         SPC&  DPC&  Acc&  S\_Acc& D\_Acc\\ \hline
         \multirow{4}*{1}&  0&  13.7&  14.9& 13.0\\
         ~&  1&  17.5&  18.0& 16.9\\
         ~&  2&  19.6&  21.1& 17.1\\
         ~&  3&  20.6&  23.1& 19.5\\ 
         \hline
         \multirow{2}*{2}&  1&  22.3&  23.0& 21.4\\
         ~&  2&  23.3&  24.1& 23.1\\\hline
    \end{tabular}}
    \caption{Test accuracies of static and dynamic group on network trained with distilled data by different SPC and DPC for MiniUCF IPC=1. Acc: test accuracies of all classes. S\_Acc: test accuracies of the static group. D\_Acc: test accuracies of the dynamic group.}
    \label{tab:video_dyn}
\end{table}

\section{Impact of Video Dynamics on Distillation}
\subsection{Selection of Static and Dynamic Group}
We utilize a pre-trained four-layer 2D convolutional network to extract features from individual frames. Subsequently, we compute the Hamming distance between the features of consecutive frames. Then, we derive the average inter-frame Hamming distance for each video segment, allowing us to ascertain the average inter-frame Hamming distance for each class.
We classify 50\% of the classes with smaller average inter-frame Hamming distances into the static group, while the remaining 50\% with larger distances are designated as the dynamic group. In Suppl.~Tab.~\ref{tab:miniucf}, we highlight the static and dynamic groups using distinct colors.
\subsection{More Detailed Results}
We show more detailed results in Suppl.~Tab.~\ref{tab:video_dyn}. We can observe that both increasing static memory and dynamic memory concurrently enhance the accuracy of static and dynamic classes. Additionally, we note that when the disparity in quantity between static memory and dynamic memory becomes larger, there is an imbalance in the accuracy of the static group and dynamic group. The reason for this lies in the random pairing of dynamic memory and static memory during sampling. If there is an excess of dynamic memory, multiple instances of dynamic memory may end up learning the same content during training, essentially augmenting the static information. Therefore, we recommend maintaining a 1:1 ratio between static memory and dynamic memory when utilizing our paradigm.

\begin{figure*}[t]
    \centering
    \includegraphics[width=0.8\linewidth]{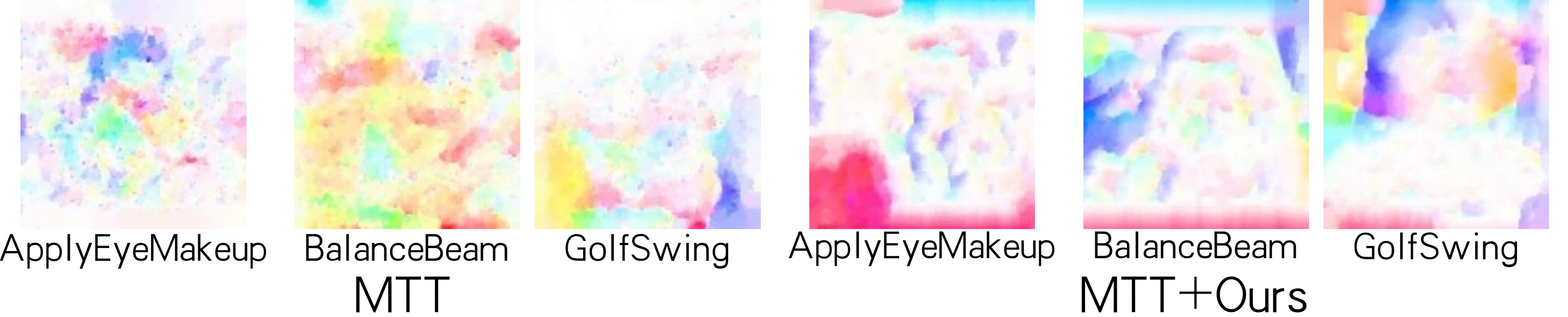}
    \caption{Optical Flows of MTT for MiniUCF IPC=1.}
    \label{fig:vis3}
\end{figure*}

\section{Visualization} 

\subsection{\added{Optical Flows}}
\added{Our distilled video data could also generalize beyond the classification task.
We show some visualized optical flow extraction following \cite{opflow} in Fig.~\ref{fig:vis3}.}

\subsection{\added{Inter-frame Differences}}
\begin{figure*}
    \centering
    \includegraphics[width=0.8\linewidth]{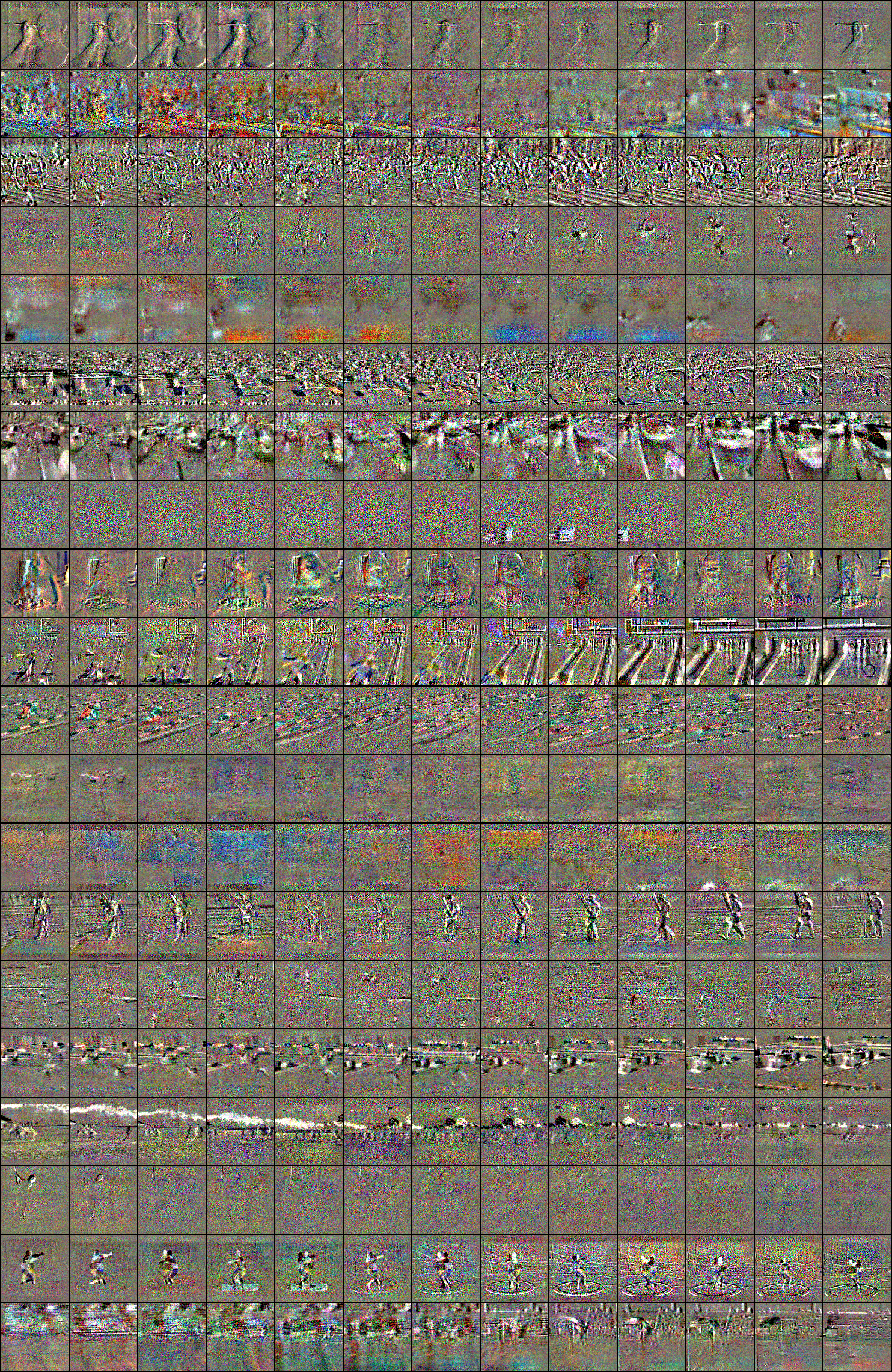}
    \caption{Inter-frame Differences of MTT for MiniUCF IPC=1.}
    \label{fig:vis1}
\end{figure*}

\begin{figure*}
    \centering
    \includegraphics[width=0.8\linewidth]{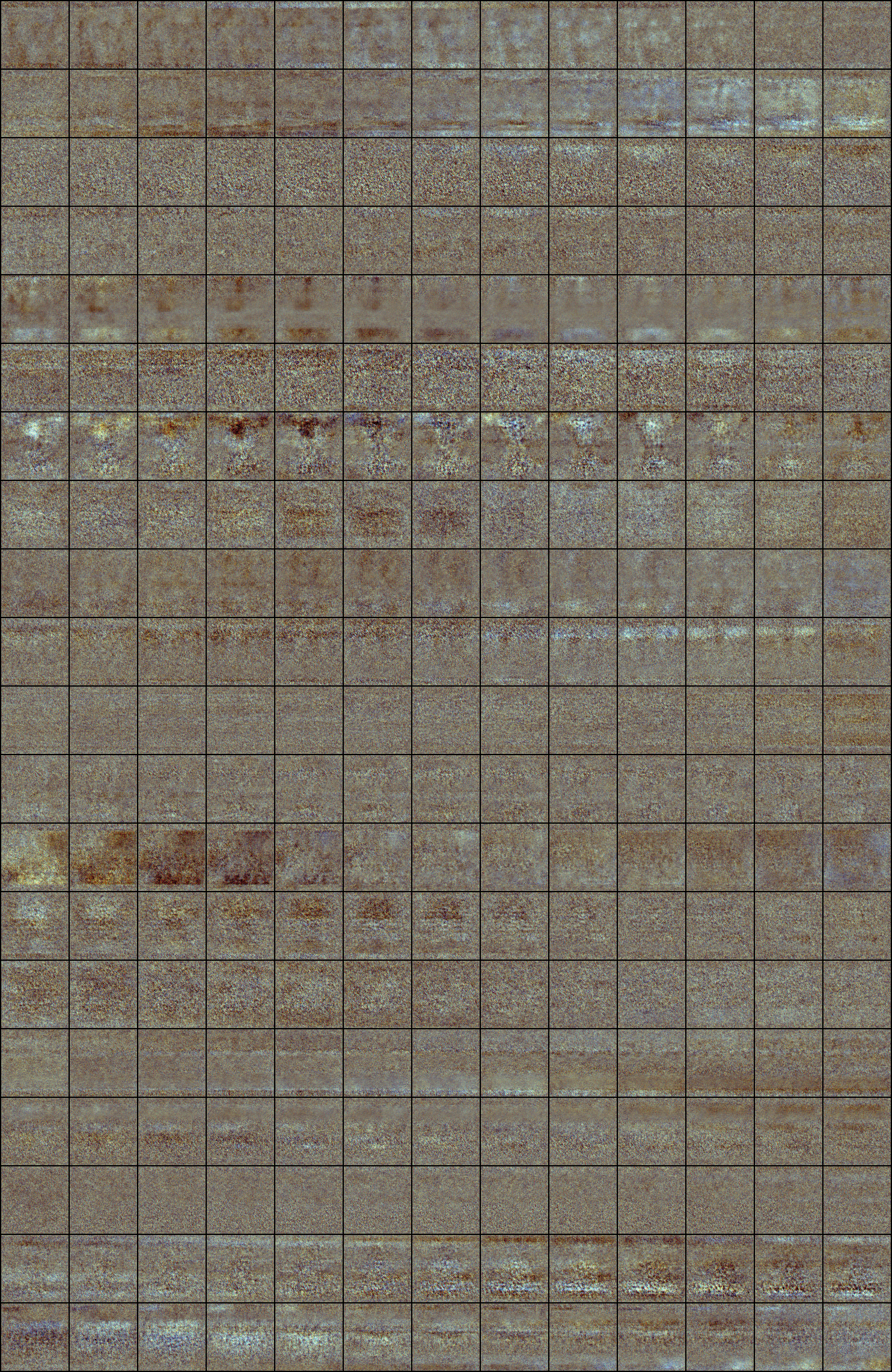}
    \caption{Inter-frame Differences of MTT+Ours for MiniUCF IPC=1.}
    \label{fig:vis2}
\end{figure*}

We show more visualized inter-frame differences of MTT~\cite{MTT} and MTT+Ours for MiniUCF IPC=1 in Fig.~\ref{fig:vis1} and Fig.~\ref{fig:vis2} (last pages).

\section{Implementation Details}

\subsection{Network Structure}
In this section, we provide a detailed introduction to models used in the experiments, including MiniC3D, CNN+GRU, and CNN+LSTM. Additionally, we compare their performance with C3D\cite{c3d} on the UCF101\cite{ucf101} and HMDB51\cite{ucf101} classification tasks.

\begin{figure}[t]
    \centering
    \includegraphics[width=1\linewidth]{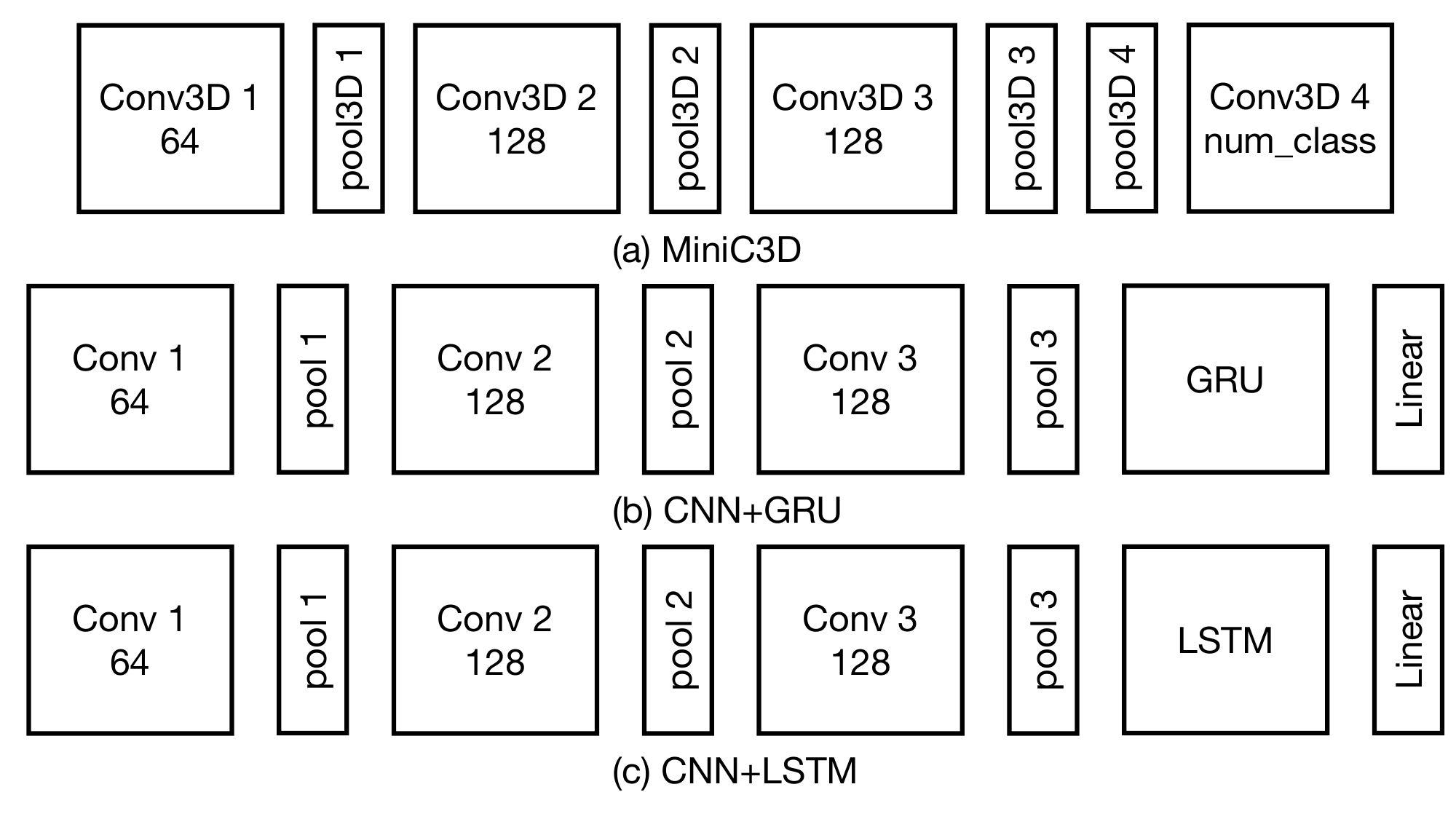}
    \caption{Structure of Models.}
    \label{fig:models}
\end{figure}

\begin{table}[t]
    \centering
    \resizebox{\linewidth}{!}{
    \begin{tabular}{c|cccc} \hline
         &  MiniC3D&  CNN+GRU&  CNN+LSTM& C3D$^\star$\cite{c3d, kinetics}\\ \hline
         UCF101~\cite{ucf101}& 33.7 & 30.4&27.92& 51.6\\
         HMDB51~\cite{hmdb51}&  28.6& 24.0& 23.2& 24.3\\\hline
    \end{tabular}}
    \caption{Top-1 Action Recognition Accuracies on Different Models. For UCF101 and HMDB51, we train and test on split 1. $\star$:~The results of C3D are referenced from \cite{kinetics}.}
    \label{tab: compare model}
\end{table}

\paragraph{MiniC3D.}
 Suppl.~Fig.~\ref{fig:models}(a) shows the structure of MiniC3D. For Conv3D blocks except for Conv3D 4, we use 3$\times $7$\times$ 7 kernels with 1$\times$ 2$\times$ 2 strides and 1$\times$ 3$\times$ 3 paddings. Conv3D 4 is used for classification, which has 1$\times $1$\times$ 1 kernel and 1$\times$ 1$\times$ 1 stride. Channels are denoted below block names in Suppl.~Fig.~\ref{fig:models}. For pooling operations, we employ a 1$\times$2$\times$ 2 kernel for max-pooling in pool3D 1, and 2$\times$ 2$\times$ 2 kernels for max-pooling in both pool3D 2 and pool3D 3. In contrast, we utilize average pooling in pool3D 4.
 
\paragraph{CNN+GRU.}
Suppl.~Fig.~\ref{fig:models}(b) shows the structure of CNN+GRU. For Conv blocks, we use 3$\times $3 kernels with 1$\times$ 1 strides and 1$\times$1 paddings. Channels are denoted below block names in Suppl.~Fig.~\ref{fig:models}. For pooling operations, we employ 2$\times$ 2 kernels for average pooling in both pool 1, pool 2, and pool 3. The GRU block is a single-layer GRU. A linear layer is used for classification.

\paragraph{CNN+LSTM.}
Suppl.~Fig.~\ref{fig:models}(c) shows the structure of CNN+LSTM. For Conv blocks, we use 3$\times$3 kernels with 1$\times$ 1 strides and 1$\times$ 1 paddings. Channels are denoted below block names in Suppl.~Fig.~\ref{fig:models}. For the pooling operations, we employ 2$\times$ 2 kernels for average pooling in both pool 1, pool 2, and pool 3. The LSTM is a single-layer LSTM. A linear layer is used for classification.

\paragraph{Comparison With Full Model.}

We show the classification results of UCF101 and HMDB51 on these models and full C3D in Suppl.~Tab.~\ref{tab: compare model}. In Suppl.~Tab.~\ref{tab: compare model}, we observe that (1) on MiniC3D, the classification accuracy of UCF101 is lower than that on full C3D; (2) on MiniC3D, the classification accuracy of HMDB51 can even exceed that on full C3D.

\begin{table}[t]
    \centering
    \begin{subtable}{.95\linewidth}
    \centering
    \resizebox{\linewidth}{!}{
        \begin{tabular}{lc|ccc} \hline
             Dataset&  IPC&  lr\_img  &batch\_syn&  syn\_steps \\ \hline
             \multirow{2}*{MiniUCF}&  1&  1e5  &50&  10\\
             ~&  5&   1e5 &128&  5\\
             \multirow{2}*{HMDB51}&  1&   1e4 &51&  10\\
             ~&  5&   1e6 &128&  5\\
             \multirow{2}*{Kinetics400}&  1&   5e5 &256&  10\\
             ~&  5&   1e7 &256&  5\\
             \multirow{2}*{SSv2}&  1&   1e5 &256&  10\\
             ~&  5&   1e6 &256&  5\\\hline
        \end{tabular}
    }
    \caption{MTT~\cite{MTT}}
    \end{subtable}
    
    \begin{subtable}{.95\linewidth}
    \centering
    \resizebox{\linewidth}{!}{
        \begin{tabular}{lc|cccc} \hline
             Dataset&  IPC&  lr\_dynamic&lr\_hal  &batch\_syn & syn\_steps \\ \hline
             \multirow{2}*{MiniUCF}&  1&   1e4&1e-3  &50&  10  \\
             ~&  5&   1e4& 1e-3 &128&  5  \\
             \multirow{2}*{HMDB51}&  1&   1e5& 1e-2 &51&  10  \\
             ~&  5&   1e6& 1e-2 &128 &  5  \\
             \multirow{2}*{Kinetics400}&  1&   1e3& 1e-2 &256&  10  \\
             ~&  5&   1e7& 1e-2 &256 &  5\\
             \multirow{2}*{SSv2}&  1&   1e4& 1e-2 &256&  10  \\
             ~&  5&   1e5& 1e-2 &256 &  5\\\hline
        \end{tabular}
    }
    \caption{MTT+Ours}
    \end{subtable}
    \caption{Hyper-parameters for MTT and MTT+Ours.}
    \label{tab:hyper-MTT}
\end{table}

\subsection{Details of Temporal Analysis}

The temporal analysis experiments in Sec.~3.3 are conducted with DM~\cite{DM} algorithm and CNN+GRU model as detailed before. 16 frames are sampled from each video with temporal stride 12, and we set the target synthetic video length also 16. For a fair comparison of time and space complexity, all the experiments are run on one NVIDIA V100 GPU (16GB), 8 cores of Intel Xeon 5218 CPU, and 20 GB memory.
The learning rate for synthetic images is set to 1.0 and that for network updating is set to 0.01. The models are trained for 10,000 iterations with a real batch size of 64, which is observed as enough for the training convergence in our experiments.

\begin{table}[t]
    \centering
    \begin{subtable}{.75\linewidth}
    \centering
    \resizebox{\linewidth}{!}{
        \begin{tabular}{lc|cc} \hline
             Dataset&  IPC&  lr\_img&  batch\_real\\ \hline
             \multirow{2}*{MiniUCF}&  1&  30&  64\\
             ~&  5&  100&  64\\
             \multirow{2}*{HMDB51}&  1&  30&  64\\
             ~&  5&  300&  64\\
             \multirow{2}*{Kinetics400}&  1&  100&  64\\
             ~&  5&  500&  128\\
             \multirow{2}*{SSv2}&  1&  10&  64\\
             ~&  5&  100&  128\\\hline
        \end{tabular}
    }
    \caption{DM~\cite{DM}}
    \end{subtable}
    \begin{subtable}{.95\linewidth}
    \centering
    \resizebox{\linewidth}{!}{
        \begin{tabular}{lc|ccc} \hline
             Dataset&  IPC&  lr\_dynamic&  lr\_hal&batch\_real\\ \hline
             \multirow{2}*{MiniUCF}&  1&  1e-4&   1e-5&64\\
             ~&  5&  1e3&   1e-6&64\\
             \multirow{2}*{HMDB51}&  1&  10&   1e-6&64\\
             ~&  5&  10&   1e-5&64\\
             \multirow{2}*{Kinetics400}&  1&  1&   1e-5&64\\
             ~&  5&  10&   1e-5&128\\
             \multirow{2}*{SSv2}&  1&  1& 1e-5&64\\
             ~&  5&  100& 1e-5&128\\\hline
        \end{tabular}
    }
    \caption{DM+Ours}
    \end{subtable}
    \caption{Hyper-parameters for DM and DM+Ours.}
    \label{tab:hyper-DM}
\end{table}

\begin{table}[t]
    \centering
    \begin{subtable}{.55\linewidth}
    \centering
    \resizebox{\linewidth}{!}{
        \begin{tabular}{lc|c} \hline
             Dataset&  IPC&  lr\_img\\ \hline
             \multirow{2}*{MiniUCF}&  1&  1e-3\\
             ~&  5&  1e-3\\
             \multirow{2}*{HMDB51}&  1&  1e-3\\
             ~&  5&  1e-3\\\hline
        \end{tabular}
    }
    \caption{FRePo~\cite{FRePo}}
    \end{subtable}
    \begin{subtable}{.7\linewidth}
    \centering
    \resizebox{\linewidth}{!}{
        \begin{tabular}{lc|cc} \hline
             Dataset&  IPC&  lr\_dynamic &lr\_hal  \\ \hline
             \multirow{2}*{MiniUCF}&  1&  1e-4&1e-3  \\
             ~&  5&  1e-1&1e-3\\
             \multirow{2}*{HMDB51}&  1&  1e-1&1e-3  \\
             ~&  5&  1e-2&1e-3\\\hline
        \end{tabular}
    }
    \caption{FRePo+Ours}
    \end{subtable}
    \caption{Hyper-parameters for FRePo and FRePo+Ours.}
    \label{tab:hyper-FRePo}
\end{table}

\subsection{Details of Full Experiments}
In the experiments, we initially fine-tune the parameters for methods without our paradigm (naively adapted methods) to achieve the best possible results. To ensure a fair comparison, we strive to maintain consistency in all other parameter settings, making adjustments only to the learning rates associated with the unique dynamic information and $\mathcal{H}$ in the methods with our paradigm.

\paragraph{Structure of $\mathcal{H}$.}
We employ two different $\mathcal{H}$. The simple one only has one Conv3D block with 3$\times $3$\times$ 3 kernels, 1$\times$ 1$\times$ 1 stride, and 1$\times$ 1$\times$ 1 padding, while the other one has one Conv3D block and one ConvTranspose3d block with middle channel 8. With the exception of DM+Ours for MiniUCF IPC=1, we consistently utilize the former.

\paragraph{Hyper-parameters for Distillation.}
We have thoroughly documented the parameters employed in our experiments in Suppl.~Tab.~\ref{tab:hyper-MTT}~\ref{tab:hyper-DM}~\ref{tab:hyper-FRePo}. Parameters not explicitly mentioned default to the values specified in the original implementation code. The specific meanings of all mentioned parameters are detailed below:\par
\textbf{lr\_img:} learning rate used to update distilled video.\par
\textbf{lr\_teacher:} learning rate used to train expert trajectories of real videos and training trajectories of distilled videos. We set it to 0.01 by default.\par
\textbf{batch\_syn:} number of distilled videos to match real videos at every iteration.\par
\textbf{batch\_real:} number of real videos to be matched at every iteration.\par
\textbf{syn\_steps:} steps of training trajectories of distilled videos to match expert trajectories at every iteration. \par
\textbf{lr\_dynamic:} learning rate used to update dynamic memory. \par
\textbf{lr\_hal:} learning rate used to update $\mathcal{H}$. \par
\textbf{expert\_epochs:} steps of expert trajectories to be matched at every iteration. We set it to 1 by default. \par
\textbf{max\_start\_epoch:} the max starting step of expert trajectories to be matched. We set it to 10 by default. \par
We train 30 expert trajectories for MiniUCF and HMDB51~\cite{hmdb51}, and 20 for Kinetics400~\cite{kinetics} and \added{Something-Something V2~\cite{ssv2}}.
Regarding whether to update lr\_teacher during the training process, we retain relatively better results for each task, such as in the case of MTT for MiniUCF IPC=1, where the performance without updating lr\_teacher surpasses that with updating.

\paragraph{Hyper-parameters for Evaluation.}
For evaluation on FRePo and FRePo+Ours, we set the learning rate to 1e-4 and trained for 1,000 epochs on MiniC3D. For other evaluations, we configure the learning rate to be 1e-2 and conduct training for 500 epochs on MiniC3D. In the case of the cross-architecture generalization test on CNN+GRU and CNN+LSTM, we set the learning rate to 1e-2 and trained for 100 epochs.

\end{document}